\documentclass[Afour,sageh,times]{sagej}

\usepackage{changes} 

\usepackage{moreverb,url}
\usepackage{float}
\usepackage[colorlinks,bookmarksopen,bookmarksnumbered,citecolor=red,urlcolor=red]{hyperref}

\usepackage{overpic}
\usepackage[capitalize]{cleveref}
\usepackage{amssymb}
\usepackage{pifont} % http://ctan.org/pkg/pifont
\usepackage[mathscr]{euscript}
\usepackage[normalem]{ulem}
\usepackage[inkscapelatex=false]{svg}

\usepackage{soul}

\newcommand\BibTeX{{\rmfamily B\kern-.05em \textsc{i\kern-.025em b}\kern-.08em
T\kern-.1667em\lower.7ex\hbox{E}\kern-.125emX}}

\def\volumeyear{2025}
\begin{document}

\runninghead{Nechyporenko et al.}

%Your manuscript’s title should be concise, descriptive, unambiguous, accurate, and reflect the precise contents of the manuscript. A descriptive title that includes the topic of the manuscript makes an article more findable in the major indexing services.
\title{\textsc{MorphIt}: Flexible Spherical Approximation of Robot Morphology for Representation-driven Adaptation}

\author{Nataliya Nechyporenko\affilnum{1}, Yutong Zhang\affilnum{1}, Sean Campbell\affilnum{1}, and Alessandro Roncone\affilnum{1}}

\affiliation{\affilnum{1} Human Interaction and RObotics [HIRO] Group, 
Department of Computer Science,
College of Engineering and Applied Science,
University of Colorado Boulder, Boulder, CO, USA\\
%\affilnum{2}SAGE Publications Ltd, UK}
}

\corrauth{Nataliya Nechyporenko,
430 UCB, 1111 Engineering Drive,
Boulder, CO, 80309, USA
}

\email{nataliya@colorado.edu}

%Abstract Please include a structured abstract of 250 words between the title and main body of your manuscript that concisely states the purpose of the research, major findings, and conclusions. If your research includes clinical trials, the trial registry name and URL, and registration number must be included at the end of the abstract. Submissions that do not meet this requirement will not be considered. For clinical trials, the trial registry name and URL, and registration number must be included at the end of the abstract. This journal includes translated abstracts, video abstracts, and plain language summaries. For more information on how to prepare a plain language summary, please see this page.

% \item[(iv)] The abstract should be capable of standing by itself,
% in the absence of the body of the article and of the bibliography.
% Therefore, it must not contain any reference citations.

% We are currently at 208 words as of april 8
\begin{abstract}
What if a robot could rethink its own morphological representation to better meet the demands of diverse tasks? Most robotic systems today treat their physical form as a fixed constraint rather than an adaptive resource, forcing the same rigid geometric representation to serve applications with vastly different computational and precision requirements. 
We introduce \textsc{MorphIt}, a novel spherical approximation framework that treats morphological representation as a tunable resource.
\textsc{MorphIt} enables automatic task-driven morphological adaptation through gradient-based optimization with tunable parameters that provide explicit control over the accuracy-efficiency tradeoff. Unlike existing approaches that rely on either labor-intensive manual specification or inflexible computational methods optimized for visualization rather than robotics, \textsc{MorphIt} generates spherical approximations up to 100x faster than optimization-based methods while maintaining superior geometric fidelity. Quantitative evaluations demonstrate that \textsc{MorphIt} outperforms baseline approaches and achieves better mesh approximation with fewer spheres. Through seamless integration with existing robotics infrastructure, \textsc{MorphIt} enables enhanced capabilities in collision detection accuracy, contact-rich interaction simulation, and navigation through confined spaces.
By dynamically adapting geometric representations to task requirements, robots can now exploit their physical embodiment as an active resource rather than an inflexible parameter, opening new frontiers for manipulation in environments where physical form must continuously balance precision with computational tractability. 

% I am not seeing other people adding websites to their abstract so removing it. 
% Website: \url{https://nataliya.dev/morphit}
\end{abstract}


\maketitle

\section{Introduction}\label{sec:intro}

% \hl{highlighted text}
% \added{new text here}         % highlights additions in green
% \deleted{old text here}       % strikes through deletions in red
% \replaced{new text}{old text} % combination of both

Biological systems possess a remarkable ability to build neural representations of the body that extend beyond physical boundaries. These morphological representations exhibit context-dependent plasticity, expanding to increase safety margins when approaching unknown entities and contracting to facilitate close-proximity manipulation \citep{bufacchi2018action, bertoni2025computational}. This adaptive quality reflects a fundamental principle: representations of physical morphology can and should adjust dynamically to meet the demands of different contexts and tasks.

\begin{figure}[t]
    \begin{overpic}[width=0.99\linewidth,percent, trim=0 0 0 20pt, clip]{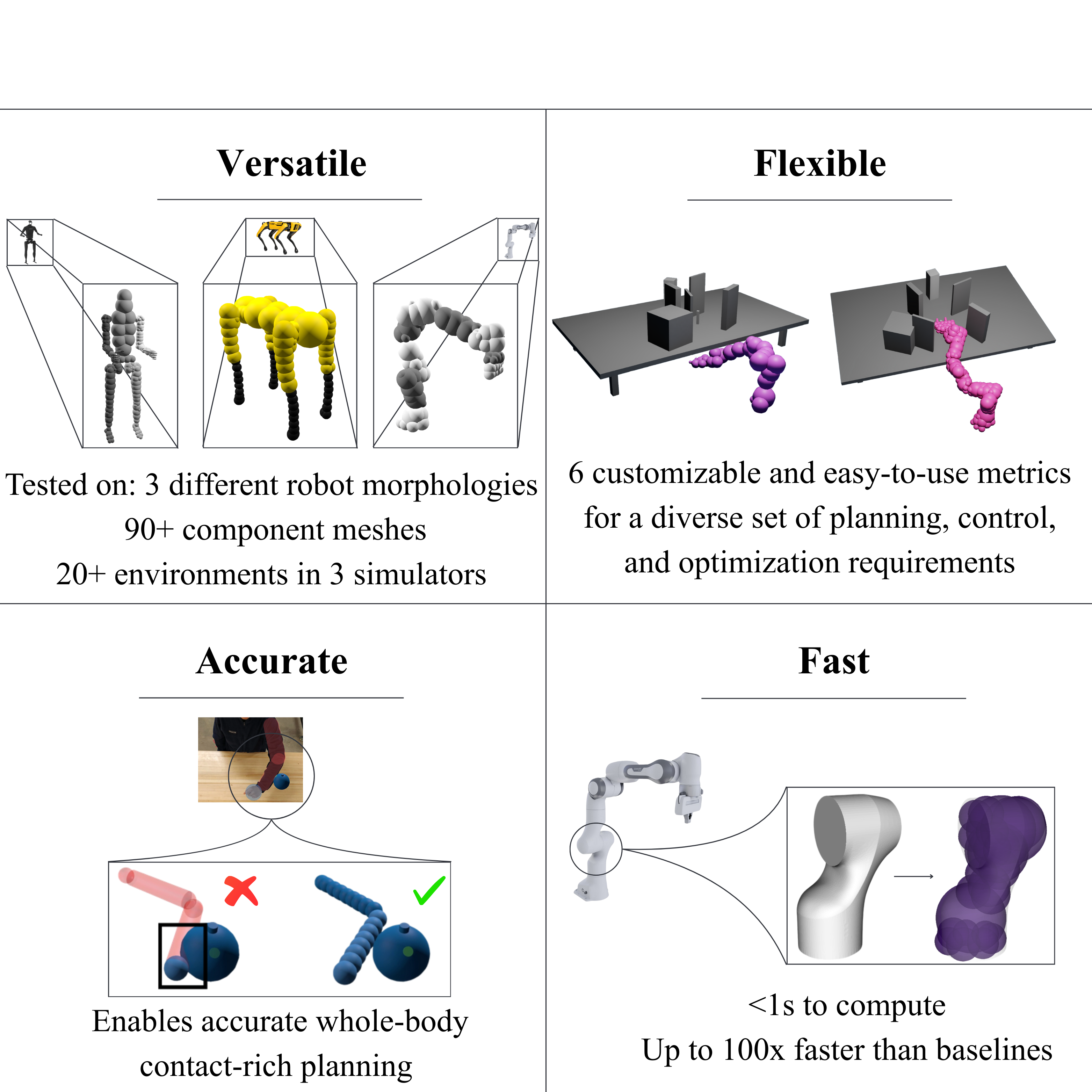}
    \put(34, 91){\huge \textsc{MorphIt}}
    \end{overpic}
    \caption {\textsc{MorphIt} enables adaptive robot morphology through four key capabilities:
    \textsl{versatile} applicability across multiple robot morphologies, meshes, and simulation environments,
    \textsl{flexible} approximation fidelity that adapts to task-specific requirements, 
    \textsl{accurate} geometric representation that improves contact-rich planning success,
    and
    \textsl{fast} computation for efficient approximation.
    }
    \label{fig:fig1}
\end{figure}

Robots also maintain internal representations of their bodies, but unlike biological systems, these representations remain rigid and fixed.
This rigidity constitutes a fundamental bottleneck on the capabilities of robotic systems, preventing robots from making full use of what their embodiment enables. Consider the tension between safe operation and dexterous manipulation: while the former requires robots to react quickly to avoid collisions in dynamic environments, the latter demands high resolution geometric representations that accurately capture the robot's physical form.
A single fixed representation cannot satisfy both requirements simultaneously.
When robots use detailed models for collision checking, distance computations that consume as much as 71\% of total runtime in motion planning pipelines become a critical bottleneck, dramatically slowing reaction times and compromising safety \citep{LaValle_2006, yu2021reducing}. 
Even with advances in GPU-based parallel processing \citep{sundaralingam2023curobo} and SIMD vector operations \citep{thomason2024motions}, these computations remain intractable for real-time control at the frequencies required in dynamic environments. 
Conversely, coarse approximations enable faster processing but introduce substantial geometric errors, leading to unwanted collisions, restricted workspace access, and severely limited manipulation capabilities.
This example demonstrates that without the ability to adapt morphological representations to match task-specific demands, robots face fundamental performance tradeoffs, unlike biological systems that fluidly transition between tasks while managing the cognitive load of body representation.

In this work, we aim to address the representational bottleneck by enabling robots to dynamically configure their morphological representations to meet the demands of different contexts and tasks. While many geometric representations exist for robot morphology, including bounding volume hierarchies \citep{dinas2015literature}, signed distance fields \citep{yang2024robotsdf}, and neural radiance fields \citep{wang2024nerf}, spherical representations have persisted in robotics applications due to their superior computational properties \citep{sundaralingam2023curobo}.
Sphere-to-sphere distance calculations have $\mathcal{O}(1)$ complexity, are analytically differentiable, inherently parallelizable, and rotationally invariant. These properties result in exceptional performance, including motion generation in 50 ms, collision-free planning at over 7000 queries per second, and contact-implicit control at 100 Hz for complex bimanual tasks \citep{sundaralingam2023curobo, kurtz2023inverse}. 
However, while these computational advantages make spheres a powerful primitive for robot representation, existing approximation methods face significant barriers to practical adoption.
Current approaches rely on either manual specification \citep{sundaralingam2023curobo, hsu2020autocomplete}, which is labor-intensive and cannot be replicated across different task requirements, or computational methods from computer graphics that optimize for visualization objectives rather than robot control and interaction \citep{wang2006variational, bradshaw2004adaptive}.
As a result, existing spherical approximation techniques fail to produce representations that are both physically meaningful and practically actionable for modern robotics.

We introduce a framework that exploits the computational advantages of spherical representations while providing the previously missing flexibility required for diverse robotic tasks. \textsc{MorphIt} automatically generates spherical approximations with tunable control over geometric accuracy and computational efficiency, enabling representations tailored to specific task demands and allowing robots to fully leverage their physical embodiment.
Our contributions are: 

\begin{enumerate}
    \item \textsc{MorphIt}, a novel method that automatically approximates robot morphology using spherical primitives (see \cref{fig:fig1} for key capabilities). By generating standard Universal Robot Description Format (URDF) files, \textsc{MorphIt} integrates seamlessly with existing robotics simulators and planning frameworks across diverse robot platforms. The complete implementation is available open source.
    
    \item Our comprehensive evaluation framework unifies geometric fidelity measures with robot-specific performance metrics, revealing that \textsc{MorphIt} generates approximations up to two orders of magnitude faster than baseline methods. Beyond speed, \textsc{MorphIt} enables task-configurable specialization: when optimized for volume coverage, it produces conservative approximations that satisfy tight collision constraints for safe obstacle avoidance, while surface-focused configurations capture fine geometric details necessary for accurate contact interactions.

    \item Through experimental validation, we demonstrate that configuring morphological representations to task requirements yields substantial performance gains. In motion planning scenarios, task-adapted representations achieve 12.2\% higher success rates while simultaneously reducing computational overhead through 62.5\% fewer spheres resulting in 65.6\% faster planning times. We show that \textsc{MorphIt} integrates easily with existing planners to enable obstacle avoidance and contact-rich manipulation strategies.
\end{enumerate}

\textsc{MorphIt} represents an important step toward task-aware morphological reasoning. We envision a future where robots reason about and strategically leverage their physical form as an active component in planning and interaction tasks, redefining morphology as a tool for physical intelligence---not just a representational limitation.
\section{Related Work}\label{sec:related}

\subsection{Representing robot morphology} \label{sec:related-A}
Geometric representations for robots and their environments include bounding volume hierarchies (BVH) \citep{dinas2015literature}, such as AABBs \citep{bergen1997efficient}, OBBs \citep{gottschalk1996obbtree}, and k-DOPs \citep{klosowski1998efficient}, which offer refined geometric fidelity but require slow hierarchical traversal. Alternative approaches include Signed Distance Fields (SDFs) with useful gradient properties but significant precomputation costs \citep{fuhrmann2003distance, macklin2020local}, voxel grids offering constant-time queries with memory inefficiency \citep{antao2019voxel, cong2021comprehensive}, and octrees requiring reinitializing with robot movement \citep{hamada1996octree, hornung2013octomap}. Point clouds struggle with surface reconstruction \citep{antao2019voxel, kim2011urban}, while mesh-based methods require iterative optimization \citep{montaut2024gjk}. Gaussian splatting lacks volumetric representation and demands numerous viewpoints for updates \citep{zhu20243d}. Neural network approaches that implicitly learn geometry \citep{yang2024robotsdf, park2019deepsdf, wang20243d, driess2022learning, wang2024nerf} provide adaptable resolution but slower inference and cannot easily integrate with existing planners (e.g. OMPL \cite{sucan2012the-open-motion-planning-library}) and simulators (e.g. Isaac Sim \cite{liang2018gpu}) 
built on standard collision engines.

\begin{figure}
    \begin{overpic}[width=0.49\textwidth, trim=0 0 0 0, clip]{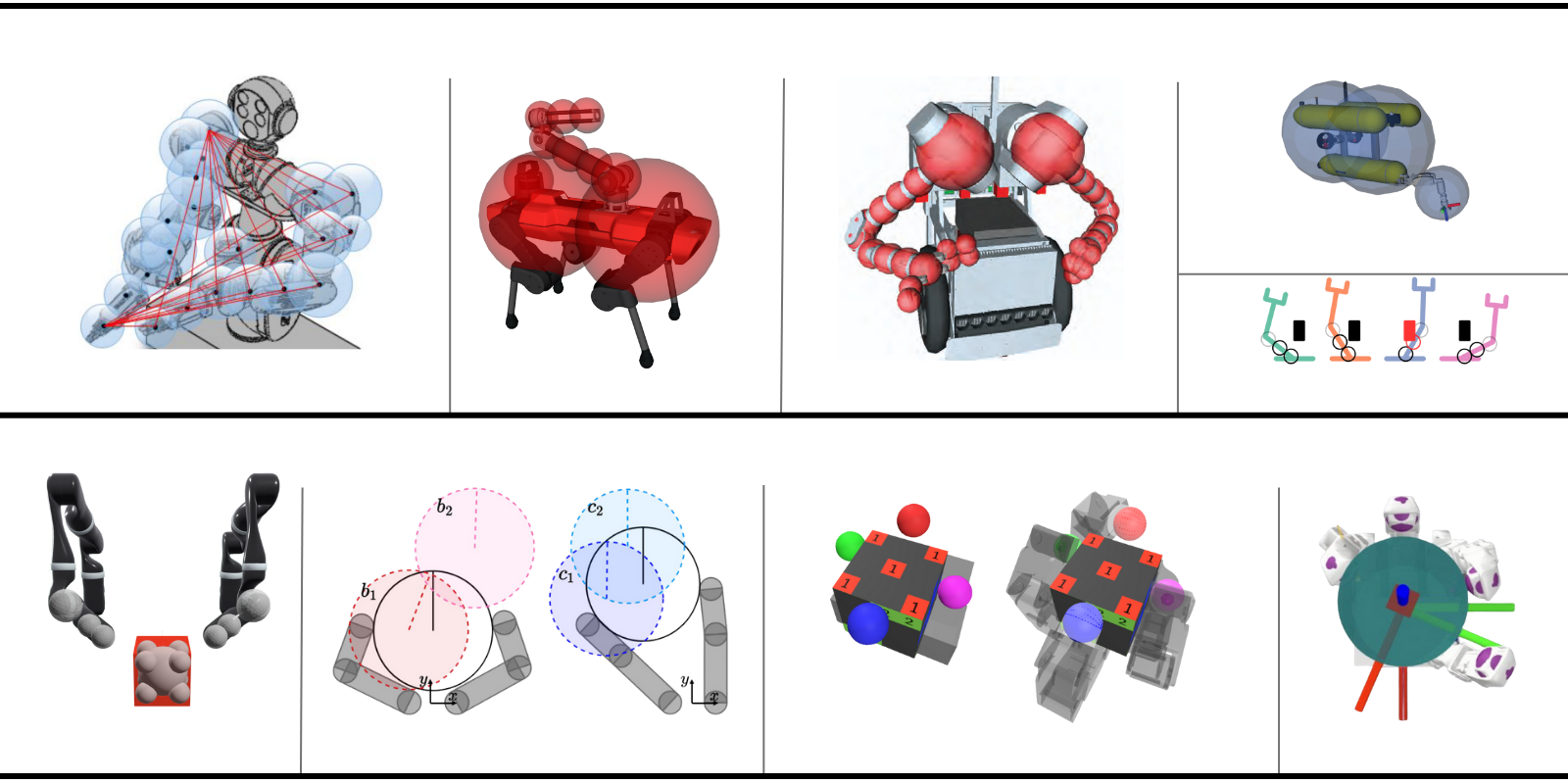}
        \put(8.0,25){\tiny \cite{lei2020real}}
        \put(32,25){\tiny \cite{chiu2022collision}}
        \put(55,25){\tiny \cite{zucker2013chomp}}
        \put(80,33.5){\tiny \cite{simoni2018novel}}
        \put(78,24.5){\tiny \cite{thomason2024motions}}
        \put(2.5,1.5){\tiny \cite{kurtz2023inverse}}
        \put(28,1.5){\tiny \cite{pang2023global}}
        \put(58,1.5){\tiny \cite{suh2025dexterous}}
        \put(84,1.5){\tiny \cite{jiang2025robust}}
        \put(5,46){\small Overly conservative approximations for obstacle avoidance}
        \put(15,20){\small Sparse approximations for contact-rich planning}
    \end{overpic}
    \caption{Current spherical approximation approaches in robotics suffer from significant limitations that restrict robot capabilities, as demonstrated by these representative examples from recent literature. Top row: Overly conservative approximations commonly used for obstacle avoidance extend well beyond actual robot geometry.
    Bottom row: Sparse approximations used in contact-rich planning focus predominantly on end-effector contact points, leaving much of the robot body unmodeled. 
    The prevalence of these suboptimal representations across diverse robot types and applications underscores the critical need for more sophisticated and adaptable approximation approaches.}
    \label{fig:related}
\end{figure}

Among various geometric representations, primitive shapes offer fast differentiable distance queries, memory efficiency, scalability, and the ability to reconstruct complex volumes and surfaces. While prior work has explored a variety of primitives, such as capsules and cubes \citep{martinez2000hierarchy, safeea2018efficient, ghosh2020fast},
spheres offer added efficiency because they are rotationally 
invariant (see \cref{sec:appendix-primitives} for an 
empirical comparison against capsules and OBBs).
Rotational invariance is key for reducing the number of optimization variables by eliminating rotational parameters and enabling simpler distance computations.
State-of-the-art motion planning systems exploit sphere-based representations to achieve the fastest planning times across hardware platforms: cuRobo \citep{sundaralingam2023curobo} achieves 60$\times$ GPU-accelerated speedup over mesh-based methods, while VAMP \citep{thomason2024motions} demonstrates 500$\times$ speedup over standard motion planning libraries on CPU.
However, their benchmarked sphere-based representations are created via labor-intensive manual packing. Developing algorithmic approaches to automatically generate effective spherical approximations with similar (or better) performance benefits remains an open and challenging problem in the field.

\subsection{Spherical approximation of 3D models} 

To understand the state of automatic generation methods, we examine two distinct research trajectories. The first trajectory originates from 
sphere packing problems, which involve arranging geometric objects within bounded regions while satisfying non-overlap constraints. This represent a well-established class of NP-hard combinatorial optimization problems with applications across manufacturing, logistics, and materials science \citep{sutou2002global, hifi2009literature}. 
Such packing formulations have been extensively applied to physical simulation in large-scale particle analysis, focusing on high-density non-overlapping spheres that prioritize physical accuracy and realistic packing densities \citep{jerier2009geometric, labra2009high, jerier2010packing, weller2010protosphere, teuber2013fast, meissenhelter2023adaptive}.
While we share the requirement for computational efficiency, we focus on overlapping spheres of different sizes, prioritizing shape representation over physical particle interaction accuracy.

The second trajectory, pioneered by the computer graphics community, emerged from real-time rendering requirements in interactive applications.
O'Rourke and Badler fit spheres to polyhedra by anchoring them to surface points and shrinking until they fit inside \citep{o1979decomposition}, later extending this to two-level hierarchies \citep{badler1979spherical}. Hubbard et al. improved scalability using medial-axis surfaces to build deeper hierarchies \citep{hubbard1996approximating}, though their Hausdorff distance metric remained computationally costly. Bradshaw and O'Sullivan built upon Hubbard's medial axis algorithm to formulate the Adaptive Medial Axis Approximation (AMAA) \citep{bradshaw2004adaptive}, employing greedy optimization to merge neighboring sphere pairs with demonstrated improvements \citep{garcia2005comparing}. Yu et al. refined this for sharp geometric features in haptic rendering \citep{yu2014simulating}, while Coumar et al. recently applied it to robot manipulators \citep{coumar2025foam}. \textsl{We use Coumar's AMAA implementation as a baseline.}

Alternatively, Cohen et al. introduced a variational framework iteratively applying Lloyd clustering \citep{lloyd1982least} for optimal piecewise-linear approximation \citep{cohen2004variational}. Wu et al. expanded the primitive types to include spheres and cylinders \citep{wu2005structure}. Wang et al. developed Variational Sphere Set Approximation (VSSA), maintaining the variational approach but introducing an ``outside volume" metric designed for collision detection and visualization \citep{wang2006variational}. Most recently, Wang's approach has been refined specifically for human shapes \citep{wu2018variational}, but has yet to be tested in the context of robots. \textsl{We include VSSA as our second baseline.}

Computer graphics methods use continuous iterative refinement through hierarchical or greedy optimization. They are generally not fit to exploit GPU parallelism due to their sequential CPU-bound nature. The computational overhead renders systematic exploration of the accuracy-efficiency tradeoff impractical for real-world deployment, motivating the development of \textsc{MorphIt}.

\begin{figure*}[t]
    \centering
    \includegraphics[width=0.98\textwidth, trim=0 65pt 0 0, clip]{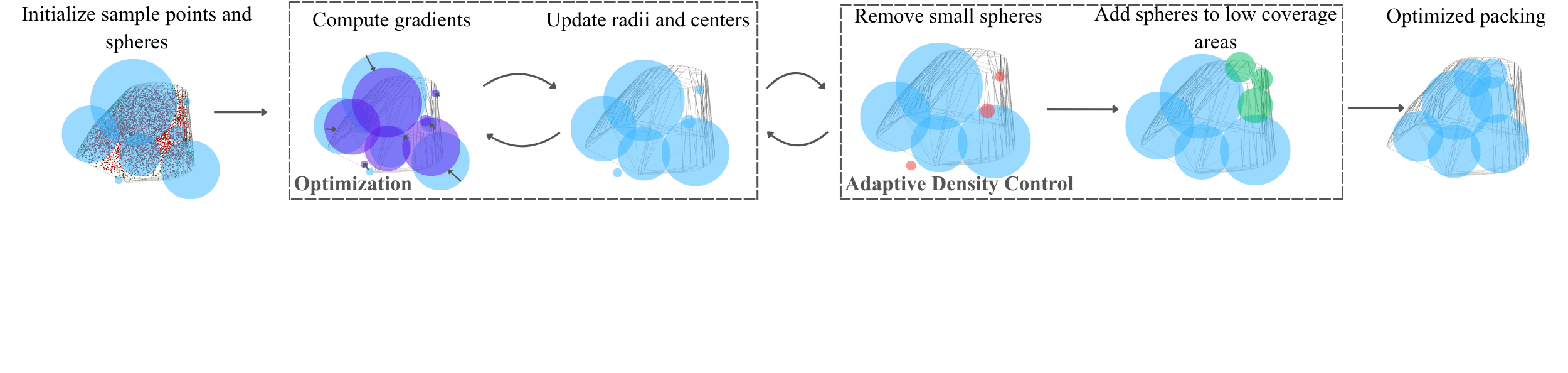} 
    \caption{Given a mesh, \textsc{MorphIt} returns a set of sphere positions and radii that represent a high fidelity packing of the original mesh. It begins by sampling points on the its surface and within its interior of the mesh. These points are used to compute surface and volume losses given sphere configurations. It then initializes sphere centers and radii according to \cref{eq:init}.
    In the main `Optimization' loop, \textsc{MorphIt} iteratively updates sphere parameters using gradient-based optimization as defined in \cref{eq:optimization}. Periodically, \textsc{MorphIt} performs `Adaptive Density Control' where it removes low-volume sphere outliers and adds new spheres (green) to uncovered regions of the mesh. 
    }
    \label{fig:algo-diagram}
\end{figure*}

\subsection{Spherical morphology in robot applications}

As seen in \cref{fig:related}, spherical approximation has been widely adopted across diverse robot morphologies including humanoids \citep{el2013optimal, lei2020real}, legged platforms \citep{chiu2022collision}, manipulators \citep{shetty2024tensor, kwon2024conformalized, carvalho2025motion}, autonomous underwater vehicles \citep{simoni2018novel}, and aerial vehicles \citep{park2020boundary}, stemming from the efficiency of sphere-based computations.

Collision checking represents a significant bottleneck, as trajectories must be discretized and each intermediate point tested for collisions \citep{pan2016fast, del1992new}. Spherical models mitigate this cost by reducing each collision query to a small number of simple sphere–sphere distance evaluations. Beyond collision checking, spherical representations enable efficient trajectory optimization through fast, analytically differentiable distance calculations. This supports smooth gradient construction in optimization-based planners \citep{zucker2013chomp, kalakrishnan2011stomp, sundaralingam2023curobo}, sampling-based planners \citep{pasricha2022pokerrt, nechyporenko2023cat, thomason2024motions}, and real-time controllers \citep{escobedo2021contact, escobedo2022framework, cai2025mpc}. 

Fast spherical distance evaluations are especially critical when robots must reason about sustained physical interaction. State-of-the-art contact-rich planning methods \citep{pang2023global, kurtz2023inverse, suh2025dexterous, jiang2025robust} rely on sphere-based models for efficient contact handling. 
However, these methods employ sparse spherical 
representations as part of their planning pipelines, typically 
concentrating spheres at the end-effector, as this makes 
contact-implicit optimization tractable (see \cref{fig:related}). 
While this is a deliberate and effective design choice, it means 
that some of the robot body remains unapproximated during planning, 
limiting the robot's ability to leverage its entire surface area 
for manipulation or provide high accuracy for whole-body obstacle 
avoidance.
We address this by developing a framework that provides flexible 
control over morphological representation, enabling richer approximations that can be adapted to.
\section{Methods}\label{sec:methods}

\begin{table*}
\caption{Definition of the six loss terms used in \textsc{MorphIt}'s composite optimization function. Cf. \cref{sec:methods:morphit}.
}
\centering
\resizebox{\textwidth}{!}{%
\begin{tabular}{|lll|}
\hline
\textbf{Loss Term}   & \textbf{Definition}                                                                                                                                                                                                     & \textbf{Description}                         \\ \hline
\textbf{\textsl{Coverage}}    & $L_{\textrm{cover}} = \mathbb{E}_{\mathbf{p} \in \textrm{interior}(\mathcal{M})}\left[\max\left(0, \min_{i}\left(\|\mathbf{p} - \mathbf{c}_i\| - r_i\right)\right)\right]$                                              & Maximizes interior volume coverage           \\ \hline
\textbf{\textsl{Overlap}}     & $L_{\textrm{overlap}} = \mathbb{E}_{i \neq j}\left[\max\left(0, r_i + r_j - \|\mathbf{c}_i - \mathbf{c}_j\|\right)\right]$                                                                                              & Minimizes overlap between spheres            \\ \hline
\textbf{\textsl{Boundary}}    & $L_{\textrm{bound}} = \mathbb{E}_{\mathbf{q} \in \textrm{surface}(\mathcal{M})}\left[\max\left(0, -\min_{i}\left(\|\mathbf{q} - \mathbf{c}_i\| - r_i\right)\right)\right]$                                              & Minimizes spheres extending outside boundary \\ \hline
\textbf{\textsl{Surface}}     & $L_{\textrm{surf}} = \mathbb{E}_{\mathbf{q} \in \textrm{surface}(\mathcal{M})}\left[\left|\min_{i}\left(\|\mathbf{q} - \mathbf{c}_i\| - r_i\right)\right|\right]$                                                       & Minimizes sphere to mesh surface distance    \\ \hline
\textbf{\textsl{Containment}} & $L_{\textrm{contain}} = \mathbb{E}_{i \neq j}\left[\max\left(0, r_j - \left(\|\mathbf{c}_i - \mathbf{c}_j\| + r_i\right)\right)^2\right]$                                                                               & Minimizes sphere within sphere volume        \\ \hline
\textbf{\textsl{SQEM}}        & $L_{\textrm{SQEM}} = \mathbb{E}_{\mathbf{q} \in \textrm{surface}(\mathcal{M})}\left[\left(\left(\mathbf{q} - \mathbf{c}_{\textrm{closest}}\right) \cdot \mathbf{n}_{\mathbf{q}} - r_{\textrm{closest}}\right)^2\right]$ & Maximizes surface reconstruction                 \\ \hline
\end{tabular}%
}
\label{tab:loss-def}
\end{table*}

Having established the need for task-adaptive and computationally efficient morphological representations, we present our framework and the baseline methods used for evaluation.

\subsection{Baseline Methods}
We compare \textsc{MorphIt} against three prior spherical approximation methods: two established optimization-based approaches from computer graphics, and one voxelization heuristic in widespread practical use.
Our first baseline, Variational Sphere Set Approximation (VSSA) \citep{wang2006variational}, minimizes the Sphere Outside Volume (SOV) metric, defined as the volume of spheres extending beyond the object's surface, using an iterative Lloyd clustering approach \citep{lloyd1982least} that partitions mesh points and generates volume-minimizing spheres for each set. VSSA's successful application to mannequin approximation for virtual clothing \citep{wu2018variational} suggests potential for articulated systems, though its effectiveness for robot morphologies is evaluated for the first time in this work.

Our second baseline, Adaptive Medial-Axis Approximation (AMAA) \citep{bradshaw2004adaptive}, constructs sphere-trees by computing a Voronoi diagram to approximate the object's medial axis, where each vertex represents a potential sphere location with radius determined by distance to the nearest surface point. AMAA adaptively refines this approximation by merging neighboring spheres or removing spheres and redistributing their coverage to neighbors, guaranteeing complete object coverage at each level of the sphere-tree hierarchy. AMAA has recently been packaged for robotic applications with enhanced mesh preprocessing \citep{coumar2025foam}. 
We note that FOAM does not introduce a new approximation algorithm but rather provides robotics-specific tooling around AMAA; our evaluation against AMAA therefore directly captures the performance of FOAM's underlying geometric method.
While both VSSA and AMAA were introduced over two decades ago, they remain relevant baselines today. The recent resurgence of these methods is driven by advances in parallel processing and GPU acceleration, which enable efficient computation of sphere-based representations at scale.

Our third baseline is the sphere approximation method from cuRoboV1~\citep{sundaralingam2023curobo}. The default Franka Panda model in this library relies on manually packed spheres. However, users can also access an API to approximate arbitrary meshes through a basic voxelization procedure. Unlike VSSA and AMAA, this method is a non-iterative heuristic with no geometric optimization objective, prioritizing computational simplicity over approximation accuracy. We include it as a baseline to contextualize the 
performance of optimization-based methods relative to the approximations 
commonly used in practice.

% https://curobo.org/get_started/2c_world_collision.html

All three baseline methods produce single-purpose 
representations motivating our work to provide flexibility over the geometry of spherical 
approximations. For complete implementation details of all methods, we refer readers to the original works and their publicly available code (see \cref{sec:appendix-code} and \cref{app:baseline_objectives}).

\subsection{Our Method: \textsc{MorphIt}}\label{sec:methods:morphit}

\textsc{MorphIt} directly optimizes a set of sphere parameters (centers and radii) through gradient-based optimization with specialized loss functions. Our approach innovates over prior works in three key ways. First, our composite cost formulation introduces six loss terms that control volume coverage, surface fidelity, and sphere interactions in an interpretable manner. By assigning tunable weights to these terms, we can generate fundamentally different approximations from the 
same mesh through interpretable, task-configurable weight settings that reflect specific task requirements. Second, our fully differentiable loss enables gradient-based optimization, which provides substantial computational speedup and efficient exploitation of GPU acceleration on hardware already standard on existing robot platforms. Third, we provide an end-to-end infrastructure that takes optimized sphere sets and exports them to standard robot description formats, enabling drop-in use across a wide range of robotics tools.

Concretely, \textsc{MorphIt} takes as input a mesh (composed of edges and vertices) representing the robot link or object to be approximated, the desired number of spheres for the representation, and weight parameters that control the relative importance of different approximation criteria (cf. \cref{eq:optimization}). The algorithm (see \cref{fig:algo-diagram}) outputs a set of spheres defined by centers and radii that approximate the input mesh according to the weighted objective function. This sphere set is then automatically converted into a Universal Robot Description Format (URDF) model that respects the robot's kinematic structure. The resulting URDF can then be used across a variety of robots and simulation environments including Isaac Sim \citep{liang2018gpu}, Pinocchio \citep{carpentier2019pinocchio}, Drake \citep{drake}, PyBullet \citep{coumans2019}, and MuJoCo \citep{todorov2012mujoco}, enabling seamless integration with existing robotics workflows.

\subsubsection{Initialization}
The model begins by sampling sphere centers $\{c_i\}_{i=1}^N$ from within the mesh volume. Sphere radii $\{r_i\}_{i=1}^N$ are initialized using a log-normal distribution that preserves the target volume given the mean radius $\bar{r}$:

\begin{equation} \label{eq:init}
    \bar{r} = \left(\frac{3 \cdot V_{\mathcal{M}}}{4N\pi}\right)^{1/3}
\end{equation}

where $V_{\mathcal{M}}$ is the mesh volume and $N$ is the desired number of spheres. This non-uniform initialization creates a natural variation in sphere sizes while maintaining the total volume, providing a better starting point for optimization than uniform radii. To enable efficient loss calculation, the algorithm pre-samples points $p \in \text{interior}(\mathcal{M})$ inside the mesh and points $q \in \text{surface}(\mathcal{M})$ on the surface for use in the subsequent loss computations.

\subsubsection{Optimization} \label{sec:methods-opt}
Our goal is to find optimal sphere parameters that approximate the target mesh $\mathcal{M}$ by minimizing a composite loss function:
\begin{equation}
\begin{split}
L_{\textrm{total}} &= w_c L_{\textrm{cover}} + w_o L_{\textrm{overlap}} + w_b L_{\textrm{bound}} \\
&+ w_s L_{\textrm{surf}} + w_t L_{\textrm{contain}} + w_q L_{\textrm{SQEM}}
\end{split}\label{eq:optimization}
\end{equation}
where $w_c$, $w_o$, $w_b$, $w_s$, $w_t$, and $w_q$ are weights for the respective loss components defined in \cref{tab:loss-def}. 
Here, the $\mathbf{c}_i$ and $r_i$ represent the center and radius of sphere $i$ (where $i = 1, \ldots, N$), while $\mathbf{p}$ and $\mathbf{q}$ denote points sampled from the interior and surface of the mesh $\mathcal{M}$, respectively. Each loss is computed as a mean over the corresponding sample points, where the $\min_i$ operation selects the closest sphere to each sample point. 

The \textsl{coverage} loss, $L_{\textrm{cover}}$, ensures spheres fill the mesh interior. The \textsl{boundary} loss, $L_{\textrm{bound}}$, prevents spheres from extending outside the mesh boundary. The \textsl{surface} loss, $L_{\textrm{surf}}$, minimizes the unsigned Euclidean distance from surface points to the nearest sphere surface. The Spherical Quadric Error Metric (\textsl{SQEM}) loss, $L_{\textrm{SQEM}}$, introduced by \cite{thiery2013}, computes the squared signed distance from surface points to their closest spheres, projected along surface normals. This normal-weighted formulation makes $L_{\textrm{SQEM}}$ more sensitive to geometric features than the directionally agnostic $L_{\textrm{surf}}$. For $L_{\textrm{SQEM}}$, $\mathbf{c}_{\textrm{closest}}$ and $r_{\textrm{closest}}$ denote the center and radius of the closest sphere to point $\mathbf{q}$, and $\mathbf{n}_{\mathbf{q}}$ represents the surface normal at $\mathbf{q}$.

Two complementary loss terms govern sphere interactions: \textsl{overlap} loss,$L_{\textrm{overlap}}$, penalizes intersecting spheres, while \textsl{containment} loss, $L_{\textrm{contain}}$, prevents one sphere from being fully enclosed within another.
Although \textsl{overlap} loss discourages sphere intersections, it admits a subtle failure mode. When sphere $j$ is fully enclosed by sphere $i$, the overlap penalty scales linearly with the radius of sphere $j$. The optimizer can therefore reduce this penalty by shrinking sphere $j$ until it becomes negligibly small and contributes nothing to the shape approximation. \textsl{Containment} loss prevents this loophole by applying a quadratic penalty proportional to the depth of penetration between the two spheres, a quantity that is independent of sphere $j$'s radius. Because shrinking sphere $j$ does not reduce and may increase this penetration depth, the optimizer cannot escape the penalty by collapsing the sphere and is instead encouraged to reposition it to cover previously uncovered regions.

\begin{figure*}
    \centering
    \includegraphics[width=1.0\textwidth, trim=0 100pt 0 0, clip]{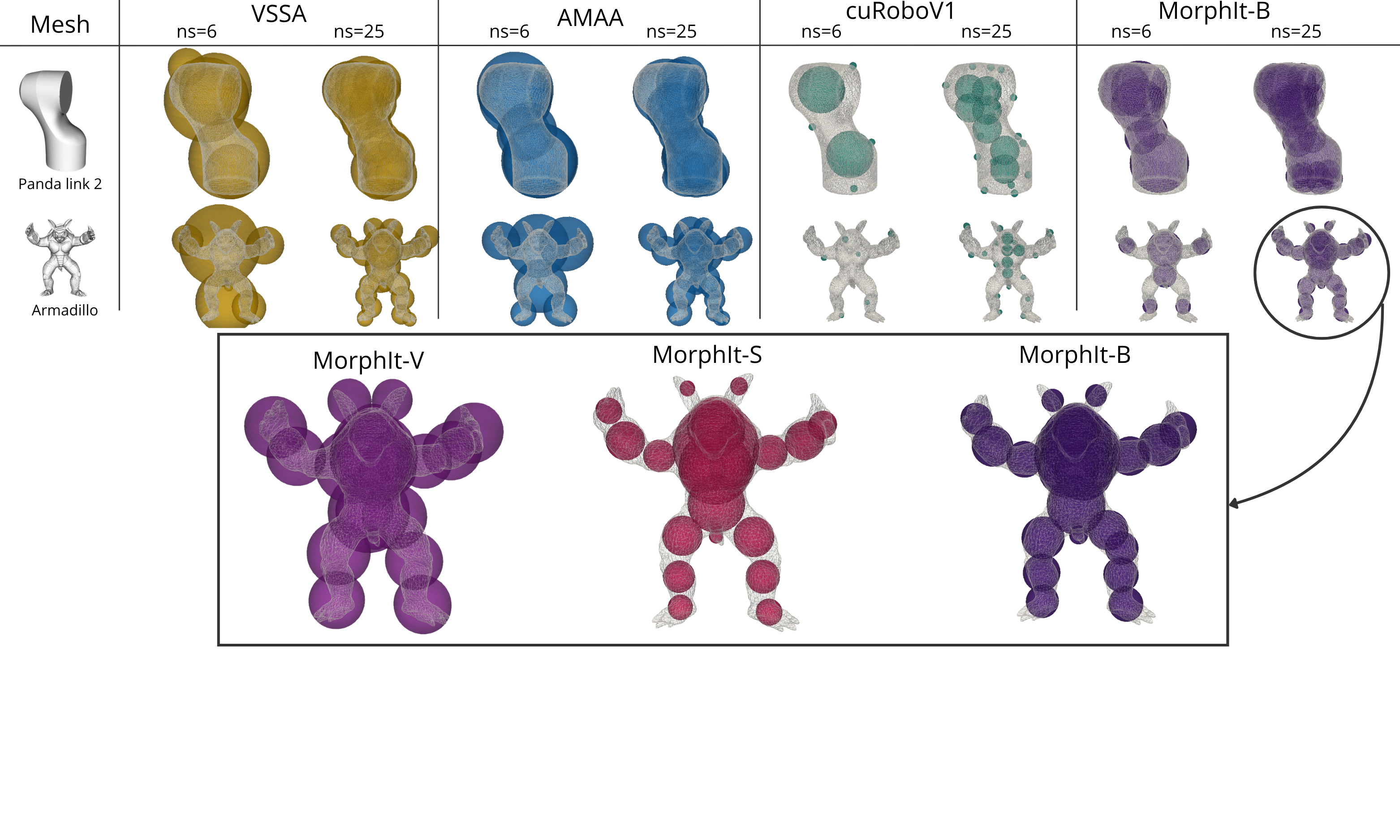} 
    \caption{Comparison of spherical approximation methods (VSSA, AMAA, cuRoboV1 and \textsc{MorphIt}) applied across different objects (Panda link 2, and Armadillo model) with varying sphere densities ($ns=6$, $ns=25$). The visualization shows how each method approaches the approximation task, with wireframe mesh overlays indicating the original geometry. The three \textsc{MorphIt} variants are compared in the 25-sphere approximation of the Armadillo model. cuRoboV1 produces visibly scattered and incomplete packings, particularly at low sphere counts, because its voxelization procedure retains only spheres whose centers clear a minimum interior distance threshold. On complex meshes with thin or concave geometry, such as the Armadillo, few or no interior points satisfy this constraint, causing the method to fall back to sparse surface sampling and leaving large portions of the volume uncovered. See \cref{app:curobo} for a full description of this procedure.}
    \label{fig:packing-ex}
\end{figure*}

Note that \textsl{containment} and \textsl{SQEM} losses use squared terms to more heavily penalize large deviations, while \textsl{coverage}, \textsl{overlap}, \textsl{boundary}, and \textsl{surface} terms use linear penalties to prioritize average-case performance. This design choice reflects the different objectives of each loss term: \textsl{containment} and \textsl{surface} reconstruction accuracy benefit from outlier sensitivity, while volume \textsl{coverage} and \textsl{boundary} constraints favor balanced distribution.

The flexibility of our method comes from the ability to select the \textsl{sphere count}, directly related to available compute resources, and the \textsl{weighing parameters}, which can be adjusted to cater to specific application requirements. For example, increasing $w_c$ encourages spheres to cover more interior volume while permitting them to extend beyond the mesh boundary, creating a padded approximation suitable for conservative collision checking. Conversely, increasing $w_b$, $w_s$, $w_q$ prioritizes accurate surface reconstruction, which is essential for contact-rich planning where small geometric deviations can significantly impact the positions and velocities of interacting objects. In \cref{sec:eval}, we evaluate several pre-tuned weight configurations to demonstrate their impact on both approximation quality and downstream robotics performance across different task scenarios.

\subsubsection{Adaptive Density Control}
Our method includes an adaptive density control mechanism that prunes ineffective spheres (small radius $r_i < r_{\textrm{threshold}}$ or centers $\mathbf{c}_i \not\in \mathcal{M}$) and adds spheres in poorly covered regions based on coverage metrics. The density control process is triggered when optimization progress plateaus, as measured by convergence metrics including loss values and gradient magnitudes. This approach, inspired by 3D Gaussian Splatting techniques \citep{kerbl20233d}, automatically adjusts sphere distribution to improve approximation quality.
In optimization terms, each density control pass is a discrete restructuring step interleaved with continuous gradient descent. This structure is analogous to basin-hopping in global optimization \citep{wales1997global}, which alternates continuous local minimization with discrete moves designed to escape the basin of attraction of a local optimum. Classical basin-hopping applies random perturbations that are accepted or rejected under a Metropolis criterion. Our restructuring step is instead deterministic and coverage-guided. It culls spheres with low coverage value and reseeds them in the most poorly covered regions of the mesh. This allows the optimizer to escape sphere arrangements that continuous gradient descent alone cannot resolve. An ablation study quantifying this contribution is provided in \cref{app:ablation}.

The training process uses PyTorch and Adam optimization with separate learning rates for centers and radii, and includes gradient clipping to ensure stable convergence. The optimization terminates after a maximum number of iterations or if the change in loss falls below a certain threshold. This fully differentiable pipeline, implemented with PyTorch tensors, enables efficient gradient computation and seamless GPU acceleration. The resulting design significantly reduces computation time and increases scalability of the approximation compared to sequential CPU-bound methods, making \textsc{MorphIt} suitable not only for rapid offline iteration but also for online morphological adaptation.
To encourage reproducibility, we open source the full 
implementation of our method (see \cref{sec:appendix-code}).

\section{Evaluation Framework}\label{sec:eval}

\subsection{Experimental Setup}

\begin{table*}
\centering
\caption{Overview of evaluation framework with three levels of metrics}
\label{tab:eval-overview}
\begin{tabular}{|l|l|c|l|c|}
\hline
\textbf{Level} & \textbf{Evaluation} & \textbf{Section} & \textbf{Quantitative Measure} & \textbf{Results} \\ \hline
\textbf{L1: Geometric} & Temporal Efficiency & \ref{sec:temp-eff} & Computation time, $t_{comp}$ (seconds) & Fig.~\ref{fig:metric-time} \\ \cline{2-5}
& Boundary Fidelity & \ref{sec:bound-fid} & Surface distance, $d_{avg}$, $d_{max}$ (mm) & Fig.~\ref{fig:metric-dist} \\ \cline{2-5}
& Spatial Occupancy & \ref{sec:spatial-occ} & Volume coverage ratios, $r_{inside}$, $r_{outside}$, $r_{union}$ & Fig.~\ref{fig:metric-vol} \\ \hline
\textbf{L2: Capability} & Collision Detection & \ref{sec:eval-colcheck} & Accuracy (\%) w.r.t. ground truth mesh & Fig.~\ref{fig:collision-accuracy} \\ \cline{2-5}
& Contact Interaction & \ref{sec:contact-short} & Object's goal state error, $p_{err}$, $r_{err}$ (mm, rad) & Tab.~\ref{tab:drake-bench} \\ \hline
\textbf{L3: Task} & Obstacle Avoidance & \ref{sec:eval-narrow} & Success (\%), Solve time t(s) &  Tab.~\ref{tab:curobo-count} \\ \cline{2-5}
& Whole-Body Manipulation & \ref{sec:eval-idto} & Object's goal state error, $\theta_{err}$ (rad) & Fig.~\ref{fig:idto} \\ \hline
\end{tabular}
\end{table*}

Having established our approach to sphere-based robot approximation, we now evaluate \textsc{MorphIt} against baseline methods (VSSA, AMAA, and cuRoboV1) across various robot meshes.
While \textsc{MorphIt} works with arbitrary meshes and robots (\cref{fig:fig1}), we focus on the Franka Emika Panda arm for quantitative evaluation. The Panda is a widely used 7-DOF manipulator in robotics research with complex link geometries that challenge approximation methods, making it an ideal benchmark. However, as seen on \cref{fig:fig1}, our approach generalizes to other robots. Furthermore, we examine the performance of our method against the benchmarks on 1--100 spheres (see \cref{sec:appendix-num-spheres} for computational details) for each of the 10 Panda meshes (spanning the 7-DOF robot), resulting in up to 1000 spheres total.
We analyze each method's optimization result for a given set of spheres, studying performance across configurations to understand the trade-offs inherent in different approximations. All experiments are repeated across multiple random seeds to account for variance arising from stochastic initialization. The number of runs and seeds used at each evaluation level is reported alongside the corresponding results.

Recall that \textsc{MorphIt} optimizes a composite objective with six loss terms controlling volume coverage, surface fidelity, and sphere interactions. The associated weights can be tuned to a specific application, but to provide practical defaults, we define three preset variants that span common robotics use cases. These presets allow users to ``dial” between configurations rather than adjusting individual weights. We define the three variants below and use them throughout our evaluation.

\textbf{\textsc{MorphIt-V}} Prioritizes interior \textsl{volume} \textsl{coverage} with weights emphasizing the \textsl{coverage} ($w_c = 4 \times 10^3$) and \textsl{containment} ($w_t = 5 \times 10^1$) losses while reducing \textsl{boundary} and \textsl{surface} losses ($w_b = 10^1$, $w_s = 10^{-1}$, $w_q = 10^2$) and minimal overlap penalty ($w_o = 10^{-1}$). This configuration generates conservative approximations suitable for collision avoidance in an uncluttered space. This variant is closest to existing methods, such as VSSA and AMAA, which primarily focus on covering the object entirely, often at the expense of precise surface fidelity.

\textbf{\textsc{MorphIt-S}} Emphasizes accurate \textsl{surface} reconstruction by increasing \textsl{boundary} and \textsl{surface} loss weights ($w_b = 5 \times 10^3$, $w_s = 10^2$, $w_q = 10^3$) while reducing volume \textsl{coverage} priority ($w_c = 10^{-2}$) and using minimal \textsl{overlap} and \textsl{containment} penalties ($w_o = 10^{-2}$, $w_t = 1$). This variant is developed for scenarios requiring precise contact modeling for manipulation tasks.

\textbf{\textsc{MorphIt-B}} Uses \textsl{balanced} weights for volume and surface reconstruction ($w_c = 10^2$, $w_b = 5$, $w_s = 5$, $w_q = 8 \times 10^2$) with moderate \textsl{overlap} and \textsl{containment} losses ($w_o = 1$, $w_t = 5$). This serves as our default configuration, providing a general-purpose approximation that balances geometric fidelity with computational efficiency across diverse robotic applications.

We distinguish between the optimization process, which is fully automatic given a weight configuration, and the weight selection process, which can take two forms. One may adopt one of the three presets above or specify any combination of the six weights directly to meet the precise demands of a given application. The presets are illustrative configurations that span common use cases, not an exhaustive set of supported behaviors. Automating weight selection via meta-learning or task-performance-based Bayesian optimization is a natural extension and is discussed in \mbox{\cref{sec:conclusion}.}

\subsection{Evaluation Methodology}

The lack of standardized evaluation criteria for spherical approximations represents a significant gap in robotics research. While morphological approximations are widely used across motion planning and control frameworks, their fidelity is rarely analyzed systematically. Existing methods in computer graphics rely on isolated metrics that favor specific objectives (e.g. VSSA relies on outside volume error while AMAA favors surface reconstruction), making cross-method comparison difficult. In robotics, systematic evaluation is even more sparse. For example, \citet{coumar2025foam} only report
simulator-specific rendering times without assessing geometric fidelity or task performance.

We address this gap by introducing a multi-level evaluation framework that assesses spherical approximations from isolated geometric properties to integrated system performance (see \cref{tab:eval-overview}). \textbf{Level 1} measures how well spheres reconstruct mesh geometry through surface distance and volume ratios, providing foundational quality measures independent of specific tasks. \textbf{Level 2} evaluates how approximation quality impacts individual robot capabilities through collision detection accuracy and contact interaction fidelity. \textbf{Level 3} assesses performance on complete robot tasks, including motion planning in cluttered environments and whole-body contact-rich manipulation. The following sections detail each metric, their qualitative significance for robots, and the quantitative values used for comparison. 

\subsection{L1 Metric: Geometric Reconstruction} \label{sec:methods-metrics}
Effective spherical approximations must accurately and efficiently capture both surface and volume reconstruction, enabling principled comparisons between methods. Our geometric reconstruction metrics are as follows:

\subsubsection{Temporal efficiency}\label{sec:temp-eff} is measured through computation time, $t_{comp}$, which is the time required to generate the sphere packing in seconds (s). 
We evaluate the computation time to approximate a mesh with spheres across different mesh complexities. 
We test performance on both \textsl{collision meshes}, simplified geometric models specifically optimized for fast collision detection calculations and \textsl{visual meshes}, higher-fidelity models with detailed surface geometry used for rendering and visualization purposes. Collision meshes typically have significantly fewer triangles to accelerate computation, whereas visual meshes preserve more geometric detail for accurate appearance. 

It is important to note that VSSA and AMAA are implemented as sequential, CPU-bound algorithms; their iterative Lloyd clustering and greedy sphere-merging procedures are not amenable to batch tensor operations and therefore cannot exploit GPU parallelism. In contrast, \textsc{MorphIt} was designed from the outset around a fully differentiable loss that is inherently parallelizable across both spheres and sample points, enabling seamless GPU acceleration via PyTorch. This implementation asymmetry is a direct consequence of the underlying algorithmic difference and should be considered when interpreting runtime comparisons. All experiments were conducted on the hardware described in \mbox{\cref{sec:appendix-code}}.

%Real-time applications require faster-than-real-time approximations to adapt to changing morphologies or task requirements.

\subsubsection{Boundary fidelity}\label{sec:bound-fid} is quantified through two surface distance metrics. Maximum distance, $d_{max}$, measures the maximum distance from any mesh surface point to the nearest sphere surface, capturing the worst-case approximation error. Average distance, $d_{avg}$, measures the average distance from mesh to sphere, providing an overall measure of surface approximation quality. Values closer to zero indicate better boundary representation.

\subsubsection{Spatial occupancy}\label{sec:spatial-occ} is evaluated through three volumetric ratios.
    Inside volume ratio, $r_{\text{inside}}$, measures the fraction of sphere volume contained within the mesh relative to the total mesh volume. Higher values indicate better coverage.
       \begin{equation}
           r_{\text{inside}} = \frac{V_{\text{spheres}} \cap V_{\text{mesh}}}{V_{\text{mesh}}}
       \end{equation}

    Outside volume ratio, $r_{\text{outside}}$,  measure the ratio of the excess sphere volume extending beyond the mesh boundary to the total mesh volume. Lower values indicate tighter approximations with minimal geometric padding.
        \begin{equation}
            r_{\text{outside}} = \frac{V_{\text{spheres}} \setminus V_{\text{mesh}}}{V_{\text{mesh}}} 
        \end{equation}

    Union volume ratio, $r_{\text{union}}$, captures the total approximation efficiency accounting for sphere overlaps. Values closer to 1 indicate optimal geometric representation.
        \begin{equation}
            r_{\text{union}} = \frac{V_{\text{spheres}} \cup V_{\text{mesh}}}{V_{\text{mesh}}} 
        \end{equation}

The union volume ratio serves as a metric for overall volume approximation quality. When $r_{\text{union}}$ diverges from the optimal value of 1, we can diagnose the source of geometric error by examining whether it originates from insufficient interior coverage ($r_{\text{inside}} < 1$) or excessive extension past the boundary of the mesh ($r_{\text{outside}} > 0$). 

Without loss of generality, we rely on Monte Carlo sampling to estimate volumes for $r_{inside}$, $r_{outside}$, and $r_{union}$. More specifically, we generate a set of random points within the bounding volume and calculate the ratio of points falling within each region (mesh, spheres, or both) to estimate the corresponding volumes. Surface metrics, $d_{max}$ and $d_{avg}$, are computed by sampling points on the mesh surface and measuring their distances to the nearest sphere.

\subsection{L2 Metric: Collision Detection} \label{sec:eval-colcheck}
Collision detection accuracy represents a critical real-world measure of how well a spherical approximation captures a robot's true geometry. 
Accurate collision detection is fundamental to safe and efficient robot operation, directly impacting a robot's ability to navigate environments without unwanted contacts.
We evaluate the impact of packing quality on robot collision checking using Pinocchio's simulation environment \citep{carpentier2019pinocchio}. 
We chose Pinocchio because it implements an accurate mesh-to-mesh collision detection algorithm, GJK++ \citep{montaut2024gjk}, which allows us to establish a reliable ground-truth baseline for actual geometric collisions against which we can compare our sphere-based approximations.
Within this environment, we first run tests with the original mesh-based Panda robot to establish a ground-truth baseline of true collision states. For each of 10 random seeds, we independently sample 100 random obstacles and a class-balanced test set of 1000 configurations (500 in collision, 500 collision-free, labeled by mesh-based ground truth). We then repeat identical scenarios, replacing the mesh robot with versions approximated by different sphere-packing algorithms, and compare the collision detection results against our ground truth data. We categorize outcomes as True Positives (TP: both sphere and mesh detect collision), True Negatives (TN: neither detects collision), False Positives (FP: sphere detects collision but mesh does not), and False Negatives (FN: mesh detects collision but sphere does not). Overall \textsl{accuracy} is measured as (TP+TN)/(TP+TN+FP+FN), representing the proportion of correct collision predictions.

\subsection{L2 Metric: Contact Interaction} \label{sec:contact-short}
Physical interaction accuracy represents another critical dimension for evaluating spherical approximations. While collision checking focuses on binary contact/no-contact outcomes, contact-rich manipulation requires precise modeling of how forces transfer between objects during sustained physical interactions. Approximation errors in these scenarios can significantly impact the resulting motion dynamics, potentially causing failures in tasks like pushing, sliding, or controlled object manipulation.

We evaluate how different sphere approximations affect the accuracy of simulated physical interactions using Drake \citep{drake}, a simulator designed for contact-rich interaction with an advanced hydroelastic contact model \citep{masterjohn2022velocity}. We design a controlled short-horizon push task in which the robot's arm makes a single lateral push contact with a cube placed on a flat surface, sweeping across it at approximately 0.2\,m/s over 1\,second. We use 100 deterministic cube placements in a tight grid are used to minimize configuration variation across runs. Errors are measured at the final settled position after 3\,seconds of simulation.
We first record joint trajectories using the original mesh-based robot model, then replay these exact joint motions using robots approximated by AMAA, VSSA, cuRoboV1, and each of the \textsc{MorphIt} variants. We quantify how accurately each spherical approximation preserves contact interaction dynamics by measuring the deviation in final object position, $p_{err}$, and orientation, $r_{err}$, compared to ground truth mesh-based simulations. By replaying identical robot trajectories with different morphological representations, we isolate the effect of geometric approximation on contact outcomes.
%[trim={left bottom right top},clip]

% \subsection{L3 Metric: Obstacle Avoidance} \label{sec:eval-narrow}
% The practical utility of spherical approximations ultimately depends on their ability to support effective motion planning in complex environments. While collision detection accuracy provides a foundation for evaluation in \cref{sec:eval-colcheck}, it is essential to assess how these approximations perform within complete planning systems under realistic constraints.
% We evaluate motion planning capabilities using CuRobo in the Isaac Sim environment \citep{liang2018gpu}, selected for its efficient trajectory optimization capabilities. Our assessment consists of two evaluations shown on \cref{fig:curobo-bench}: Evaluation A utilizes 8 environments from the benchmark by \cite{fishman2022mpinets}, with 100 variations of perturbed goal positions (maximum 20cm displacement) per environment to measure success rates and the final position error, $p_{err}$, and rotations error, $r_{err}$, of the end-effector from the goal state. Evaluation B quantifies the minimum number of spheres, $\alpha$, required for navigating a narrow passage while tracking time, $t$, to find a solution. All trajectories are validated by replaying them with high-fidelity mesh models to confirm collision avoidance.

\subsection{L3 Metric: Obstacle Avoidance} \label{sec:eval-narrow}
The practical utility of spherical approximations ultimately depends on their ability to support effective motion planning in complex environments. While collision detection accuracy provides a foundation for evaluation in \cref{sec:eval-colcheck}, it is essential to assess how these approximations perform within complete planning systems under realistic constraints.
Building on the collision detection analysis in \cref{sec:eval-colcheck}, we focus this evaluation on the three methods that achieved zero false negatives, VSSA, AMAA, and \textsc{MorphIt-V}, in order to isolate how efficiently each method can support motion planning while maintaining the strict safety guarantees required for obstacle avoidance.
We evaluate motion planning capabilities using CuRoboV1 in the Isaac Sim environment \citep{liang2018gpu}, selected for its efficient trajectory optimization capabilities. Our assessment consists of two evaluations shown on \cref{fig:curobo-bench}: Evaluation A utilizes 7 environments from the benchmark by \cite{fishman2022mpinets}, with 100 variations of perturbed goal positions (maximum 20cm displacement) per environment to measure success rate and planning time, $t$, Evaluation B quantifies the minimum number of spheres, $\alpha$, required for navigating a narrow passage while tracking time, $t$, to find a solution. All trajectories are validated by replaying them with high-fidelity mesh models to confirm collision avoidance.

\subsection{L3 Metric: Whole-body Manipulation} \label{sec:eval-idto}

In \cref{sec:contact-short}, we examine interaction errors over short time horizons, focusing on immediate contact dynamics during discrete interactions. While short-term accuracy is important, extended manipulation tasks require sustained contacts across longer sequences. However, existing contact-rich planning methods rely on sparse sphere representations (see bottom row of \cref{fig:related}) that predominantly focus on end-effector contact points, limiting whole-body manipulation capabilities.
We explore how \textsc{MorphIt} enables whole-body contact capabilities by integrating it with Inverse Dynamics Trajectory Optimization (IDTO) \citep{kurtz2023inverse}. IDTO is a contact-implicit optimization method that formulates robot motion planning as a nonlinear least-squares problem where generalized positions are the only decision variables. Given a start and a goal pose for an object, IDTO generates a trajectory for the robot to move the object to the desired position. While the contact points are generated implicitly by the algorithm, the robot representation is reduced to only a few spheres, effectively constraining contact interactions to predetermined locations on the robot geometry. We adapt (see \cref{sec:appendix-idto} for more details) the open-source version of IDTO for our evaluation.

Our comparative analysis focuses on a task where a 7-DOF Kinova Jaco arm must rotate a \textsl{spinner}, a 1-DOF sphere rotating around its axis. For this evaluation, we generate 50 random spinner placements within the workspace of the arm. Each placement is free of contact at the initial arm pose and lies within reach of the whole-body representation during the nominal arm motion, and the task requires rotating the spinner to a goal angle of $2$\,rad. Next, we use IDTO to solve the trajectory optimization problem twice for each placement, first with IDTO's original sparse three-sphere robot representation, which places spheres only at the wrist and fingertip, and then with our whole-body \textsc{MorphIt} approximation of 126 spheres covering every link. We execute each attempt in closed loop, where IDTO replans as a model predictive controller and the spinner moves only through simulated physical contact with the arm. Rather than measuring binary success rates, we calculate the angular error, $\theta_{err}$, between the goal state, $\theta_{goal}$, and the closest spinner state reached during the execution, so that $\theta_{err} = \min_t \lvert\theta_{goal} - \theta(t)\rvert$. We additionally report the fraction of trials that reach the goal angle as a function of time. By comparing these errors across the random scenarios, we assess how whole-body contact modeling impacts the optimizer's ability to complete a task.

% Our comparative analysis focuses on a task where a two-link planar arm must rotate a \textsl{spinner}, a 1-DOF sphere rotating around its axis. For this evaluation, we generate 100 random configurations of the robot and spinner. Next, we use IDTO to solve the trajectory optimization problem twice for each configuration: first with IDTO's original single-sphere robot representation, and then with our whole-body \textsc{MorphIt} approximation. Rather than measuring binary success rates, we calculate the angular error,  $\theta_{err}$, between the goal state, $\theta_{goal}$, and the final spinner position, $\theta_{final}$, after the robot's manipulation attempt. By comparing these errors across the random scenarios, we assess how whole-body contact modeling impacts the optimizer's ability to complete a task.

\section{Results \& Discussion}\label{sec:results}

\subsection{Results L1: Geometric Reconstruction} \label{sec:results-geo-rec}
\begin{figure*}
    \centering
    \includegraphics[width=0.99\textwidth]{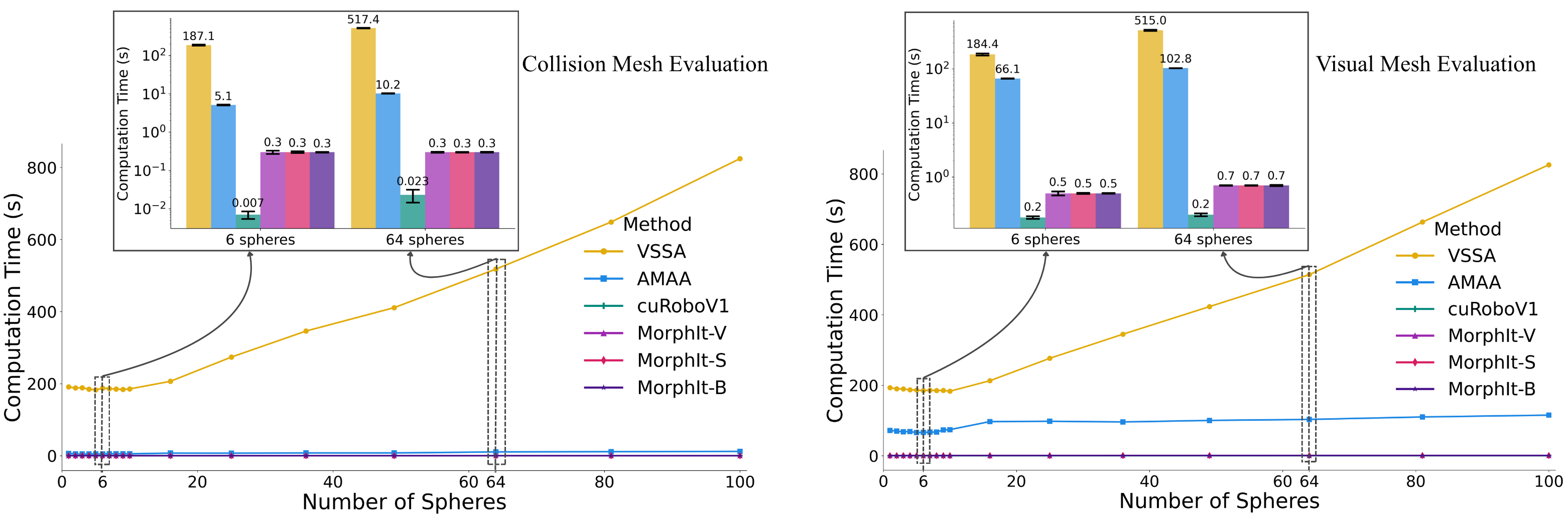}
    \caption{The impact of the number of spheres (1-100) and mesh complexity
    on the average computation time, $t_{comp}$, needed to approximate the
    ten Panda link meshes. Collision meshes (left) are simplified geometric models with approximately $10^2$ vertices and $10^3$ edges optimized for
    fast collision detection, while visual meshes (right) are high-fidelity
    models with approximately $10^4$ vertices and $10^5$ edges used for
    detailed rendering and visualization. Each data point is the mean computation time over the ten link meshes. In the bar graphs shown for the two representative sphere counts, whiskers denote $\pm 1$ std of this mean across 25 random seeds.
    }
    \label{fig:metric-time}
    \vspace{3pt}
\end{figure*}

\begin{figure*}
    \centering
    \includegraphics[width=1.0\textwidth]{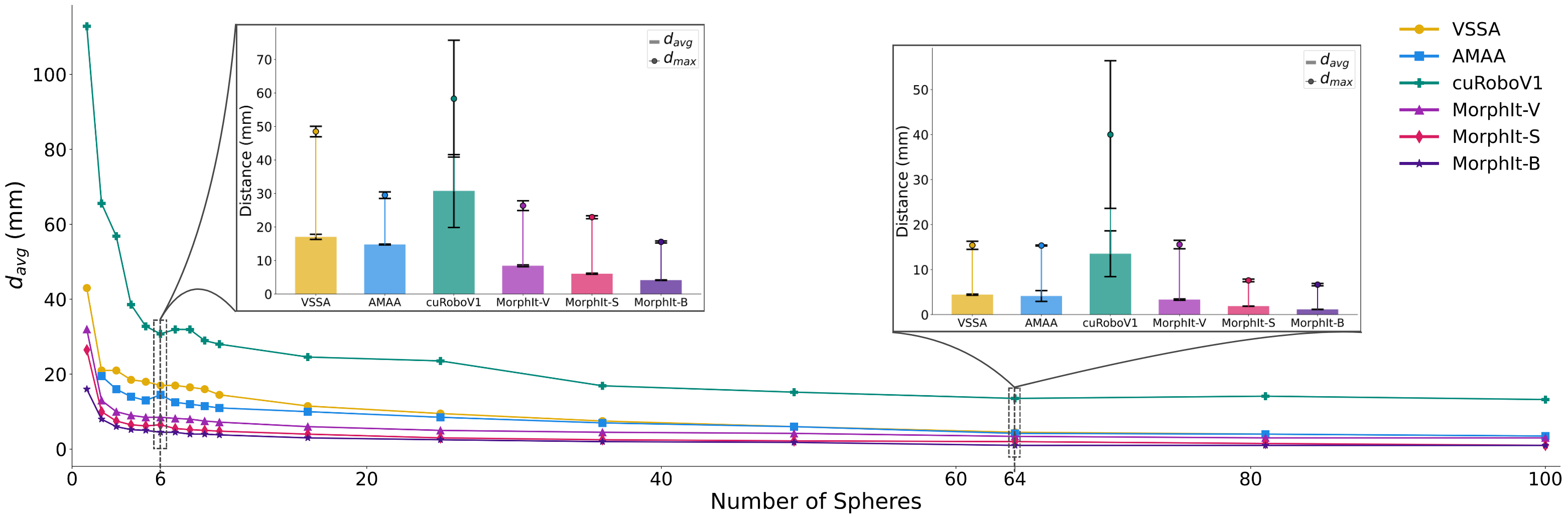} 
    \caption{The average distance, $d_{avg}$, from the mesh to sphere surface with an increasing number of spheres used for each Panda arm link. For two representative data points, we construct a bar graph to take a closer look at the average and maximum distance, $d_{max}$, between the mesh and spherical approximation surfaces. Each data point is the mean distance over the ten link meshes. In the bar graphs, whiskers on both $d_{avg}$ and $d_{max}$ denote $\pm 1$ std of this mean across 25 random seeds.}
    \label{fig:metric-dist}
    \vspace{3pt}
\end{figure*}

\begin{figure*}
    \centering
    \includegraphics[width=1.00\textwidth]{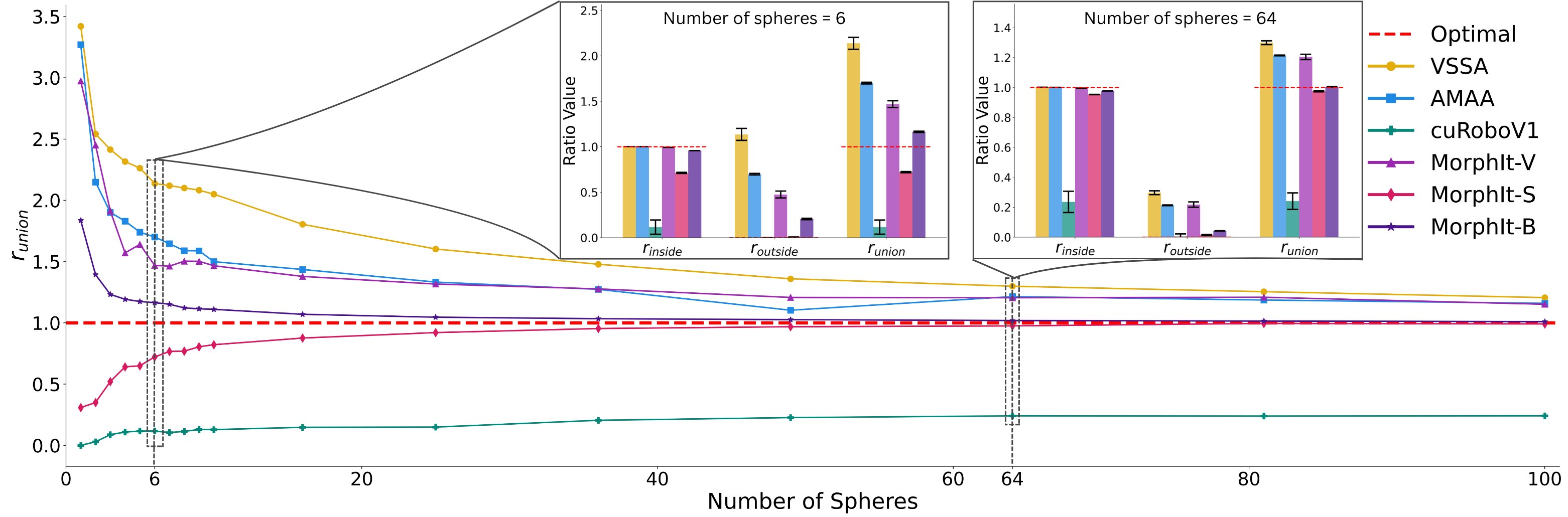} 
    \caption{The correlation between the $r_{union}$ volume ratio and an
    increasing number of spheres used for approximating each link of the
    Panda arm. For two representative data points we analyze the inside,
    $r_{inside}$, and outside, $r_{outside}$ volume ratios which inform the
    deviation from the ideal $r_{union}=1$. Each data point is the mean volume ratio over the ten link meshes. In the bar graphs, whiskers on the volume ratios denote $\pm 1$ std of this mean across 25 random seeds.}
    \label{fig:metric-vol}
    \vspace{3pt}
\end{figure*}

\begin{figure*}
    \centering
    \includegraphics[width=1.0\textwidth]{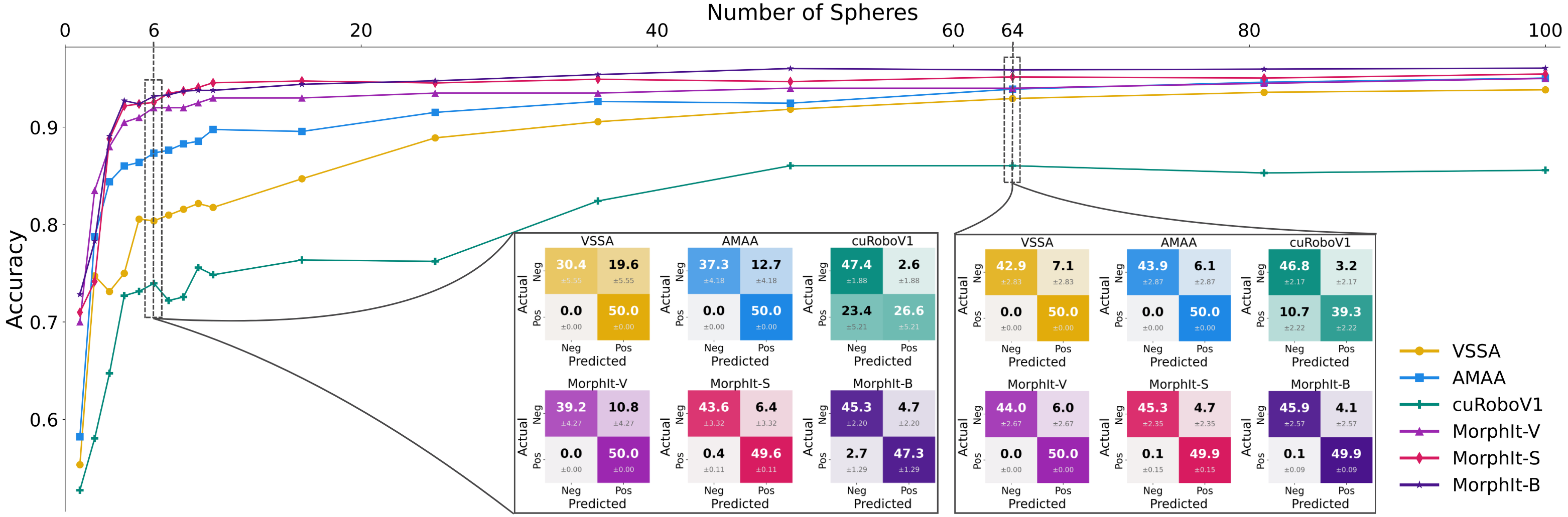} 
    \caption{Comparison of the accuracy, defined in \cref{sec:eval-colcheck}, between the approximation methods with an increasing number of spheres, $ns$, used for the approximation of each robot link. For two representative data points we show the confusion matrices demonstrating the percentage of collision detection outcomes across 1000 test cases. Accuracy is reported as mean $\pm$ 1 std across 10 random seeds. Each seed independently samples 100 obstacles and a class-balanced test set of 1000 configurations (500 in collision, 500 collision-free, labeled by mesh-based ground truth), on which all methods are evaluated.}
    \label{fig:collision-accuracy}
\end{figure*}
\subsubsection{FR3 Panda Robot Evaluation}

We evaluate \textsc{MorphIt} across multiple dimensions to demonstrate its effectiveness in robot morphology approximation. A qualitative visualization of the packing quality is shown in \cref{fig:packing-ex}. Our comprehensive analysis examines physical fidelity metrics, collision detection accuracy, contact interaction dynamics, and motion planning performance across various robotic tasks and environments.

Our performance analysis reveals significant distinctions between the evaluated methods. \cref{fig:metric-time} shows that VSSA scales well to complex meshes but struggles with larger numbers of spheres. In contrast, AMAA does not scale well to complex meshes but handles large quantities of spheres efficiently. As a non-iterative heuristic, cuRoboV1 requires the least computation time overall. However, among the optimization-based methods, all \textsc{MorphIt} variants achieve superior speed across varying sphere counts and mesh complexities. \textsc{MorphIt} generates approximations up to two orders of magnitude faster than VSSA and AMAA. This speedup reflects both algorithmic design and the GPU parallelism enabled by \textsc{MorphIt}'s differentiable formulation. By reducing computation time from minutes to sub-seconds, \textsc{MorphIt} opens a new frontier in morphological adaptation, one where robots can now adapt their geometric representations in real-time as task demands evolve. This enables previously unseen capabilities: dynamic morphological adaptation when transitioning between open spaces and clutter, instant reconfiguration when grasping objects or tools that extend the robot's effective embodiment, and on-the-fly adjustment for modular robots as their physical form changes.

While computational speed is critical, it must be matched with geometric accuracy that allows the robot to accurately represent itself in the environment. \cref{fig:metric-dist} demonstrates that \textsc{MorphIt-B} consistently outperforms its variants and the baselines for both average and maximum distance metrics across different sphere counts. While \textsc{MorphIt-V} initially outperforms the baselines for a lower number of spheres, it eventually converges to similar distances for a larger number of spheres. On the other hand, \textsc{MorphIt-S} continuously improves with an increasing number of spheres.

To better understand how these methods manage spatial representation, we also examine volume-based metrics. Since each sphere adds to the computational cost of collision checking, it is essential that every part of each sphere's volume contributes meaningfully to the mesh approximation.
\cref{fig:metric-vol} shows that while \textsc{MorphIt-V} achieves near-optimal $r_{\text{inside}}$ coverage similar to the baselines by prioritizing complete interior volume representation, \textsc{MorphIt-S} achieves minimal $r_{\text{outside}}$ extension by constraining spheres within mesh boundaries at the cost of reduced interior coverage. \textsc{MorphIt-B}'s ability to balance interior and exterior volume constraints is demonstrated by $r_{\text{union}}$, which remains closest to the ideal value of 1 across both lower and higher sphere counts. Despite its speed, cuRoboV1 exhibits the largest surface distances among all methods across all sphere counts (\cref{fig:metric-dist}), and severely under-covers the mesh interior with $r_{\text{inside}}$ approaching zero at low sphere counts (\cref{fig:metric-vol}), confirming that computational simplicity comes at a significant cost to geometric fidelity.

\subsubsection{Generalization Across Robot Morphologies} \label{sec:generalization} While our quantitative analysis focuses on the Panda for depth and coherence, \textsc{MorphIt} is not specific to any one robot. The method operates over the link-and-mesh decomposition that is standard to URDF-based robot descriptions, which makes it directly applicable to any robot that ships in this format. Table~\ref{tab:cross-robot} reports \textsc{MorphIt-B} results across five robots spanning manipulator, quadruped, and humanoid morphologies (Panda, UR5, iiwa14, Spot, and Unitree H1) using the same Panda-tuned weight configuration applied unchanged to every robot. The geometric metrics fall in a similar range across all five platforms, with $r_{\text{union}}$ between 1.18 and 1.25 at $n=6$ and converging toward 1.04--1.07 at $n=64$, and $d_{\text{avg}}$ remaining within a few millimeters of the Panda baseline. Without loss of generality, we continue to use the Panda for the remaining evaluations, but these results indicate that the same approach transfers to morphologies the method was not specifically tuned for.

\begin{table}[t]
\centering
\caption{MorphIt-B applied unchanged across five robots spanning manipulator, quadruped, and humanoid morphologies. Values are reported as mean $\pm$ 1 std across 25 random seeds,
where each seed's value is averaged over all the links of each robot.}
\label{tab:cross-robot}
\footnotesize
\setlength{\tabcolsep}{3pt}
\begin{tabular}{l c c c c}
\toprule
 & \multicolumn{2}{c}{$n=6$} & \multicolumn{2}{c}{$n=64$} \\
\cmidrule(lr){2-3} \cmidrule(lr){4-5}
Robot & $r_{\text{union}}$ & $d_{\text{avg}}$ & $r_{\text{union}}$ & $d_{\text{avg}}$ \\
\midrule
Panda  & $1.184\!\pm\!0.006$ & $4.81\!\pm\!0.08$ & $1.041\!\pm\!0.002$ & $1.50\!\pm\!0.02$ \\
Spot   & $1.217\!\pm\!0.025$ & $7.78\!\pm\!0.93$ & $1.035\!\pm\!0.000$ & $2.16\!\pm\!0.04$ \\
UR5    & $1.245\!\pm\!0.025$ & $8.05\!\pm\!0.39$ & $1.073\!\pm\!0.004$ & $2.87\!\pm\!0.02$ \\
iiwa14 & $1.190\!\pm\!0.028$ & $7.04\!\pm\!0.33$ & $1.061\!\pm\!0.005$ & $2.52\!\pm\!0.05$ \\
H1     & $1.187\!\pm\!0.003$ & $6.29\!\pm\!0.34$ & $1.036\!\pm\!0.003$ & $1.92\!\pm\!0.05$ \\
\bottomrule
\end{tabular}
\end{table}

\subsection{Results L2: Collision Detection}
Effective robot approximation requires balancing false negatives (FNs), which cause missed collisions, against false positives (FPs), which trigger unnecessary avoidance in free space. \cref{fig:collision-accuracy} illustrates how \textsc{MorphIt} navigates this tradeoff as each variant strikes a different balance optimized for specific operational requirements.

At six spheres per link, VSSA and AMAA generate over twice as many FPs as \textsc{MorphIt-B} by prioritizing volume coverage over geometric precision. cuRoboV1 exhibits the opposite failure mode: its interior voxelization method produces the highest FN rate among all methods at both sphere counts, systematically missing near-surface collisions and making it the least safe method despite its computational simplicity. \textsc{MorphIt-V} achieves zero FNs, matching VSSA and AMAA and making it ideal for safety-critical environments like hospitals where missed collisions are unacceptable. \textsc{MorphIt-S} achieves the lowest FPs, making it ideal for navigating narrow passages (\cref{fig:curobo-bench}) or delicate manipulation where false alarms cause the robot to become unnecessarily stuck. \textsc{MorphIt-B} balances these extremes for general-purpose use. While VSSA and AMAA each offer a single fixed tradeoff, \textsc{MorphIt} spans from safe-at-all-costs to precision-first configurations. Coupled with the computational efficiency demonstrated in \cref{fig:metric-time}, this tunability becomes a real-time capability that enables robots to treat their morphological representation as an adaptive resource, continuously adjusting geometric resolution to match instantaneous task demands.

\subsection{Results L2: Contact Interaction}

\begin{table}[t]
\centering
\caption{Contact interaction errors over a short-horizon push. Values are reported as mean $\pm$ 1 std across 25 seeds, evaluated over the same 100 deterministic cube placements.}
\label{tab:drake-bench}
\footnotesize
\setlength{\tabcolsep}{3pt}
\begin{tabular}{l cc cc}
\toprule
& \multicolumn{2}{c}{$n_s = 6$} & \multicolumn{2}{c}{$n_s = 64$} \\
\cmidrule(lr){2-3}\cmidrule(lr){4-5}
Method & $p_{err}$(mm) & $r_{err}$(deg) & $p_{err}$(mm) & $r_{err}$(deg) \\
\midrule
VSSA      & 43.96 $\pm$ 1.78          & 43.19 $\pm$ 3.25          & 40.93 $\pm$ 4.66                  & 30.65 $\pm$ 12.70 \\
AMAA      & 44.11 $\pm$ 2.26          & 42.95 $\pm$ 2.80          & 37.82 $\pm$ 0.47                  & 9.70 $\pm$ 1.93 \\
cuRoboV1  & 48.90 $\pm$ 0.90          & 30.55 $\pm$ 3.49          & 29.80 $\pm$ 2.54                  & 18.87 $\pm$ 3.10 \\
MorphIt-V & 43.02 $\pm$ 3.12          & 40.68 $\pm$ 5.96          & 35.52 $\pm$ 1.32                  & 23.53 $\pm$ 7.05 \\
MorphIt-S & \textbf{35.58 $\pm$ 0.83} & \textbf{18.14 $\pm$ 3.03} & \textbf{15.61 $\pm$ 0.92}         & \textbf{2.31 $\pm$ 1.69} \\
MorphIt-B & 36.32 $\pm$ 2.27          & 28.20 $\pm$ 3.74          & 26.46 $\pm$ 8.79                  & 4.91 $\pm$ 3.89 \\
\bottomrule
\end{tabular}
\end{table}

An accurate approximation must not only represent geometry well, but also preserve the dynamics of physical interaction. This is especially important in contact-rich tasks, where a robot’s behavior is driven by the interaction between itself and the environment. As shown in \cref{tab:drake-bench}, where $p_{err}$ and $r_{err}$ represent the final position and rotation errors of the object when interacting with sphere-approximated robots compared to the ground-truth mesh-based robot, the object trajectory resulting from a contact interaction with the \textsc{MorphIt-S}-based approximation most closely follows that of the ground-truth mesh. 
In the ground-truth interaction, the original mesh translates the object by 55\,mm and rotates it by approximately 40\textdegree. At a low resolution of $n_s=6$, large errors occur because they fail to push the object far enough. Since they prioritize volume, they place sparse spheres along the medial axis rather than the outer surface, creating intermittent contact that fails to transfer momentum. At a higher resolution of $n_s=64$, most methods successfully capture the correct contact direction; instead, their remaining errors stem from a failure to sustain contact-patch coverage throughout the full sweeping motion.
Unlike the other two optimization variants, \textsc{MorphIt-S}'s weights were designed to specifically optimize for surface representation, prioritizing accurate contact point modeling over volume coverage. This design choice enables \textsc{MorphIt-S} to capture the precise geometric features necessary for realistic contact dynamics, resulting in object motion that closely matches the behavior observed with the original mesh geometry.

This alignment allows \textsc{MorphIt} to simulate contact dynamics more accurately than competing methods. Such fidelity is essential for model-based approaches. Algorithms like Model Predictive Control and Reinforcement Learning rely on accurate forward predictions to plan effective actions, and even small errors in contact modeling compound over time, causing trajectories to diverge from reality and result in task failure. By preserving the geometric fidelity necessary for realistic contact dynamics, \textsc{MorphIt} enables these algorithms to learn and plan with world models that accurately capture physical reality, bridging the simulation-to-reality gap that hinders contact-rich manipulation.

\subsection{Results L3: Obstacle Avoidance}
\begin{figure}[t]
 \begin{overpic}[width=0.48\textwidth, trim=0 40pt 0 0, clip]{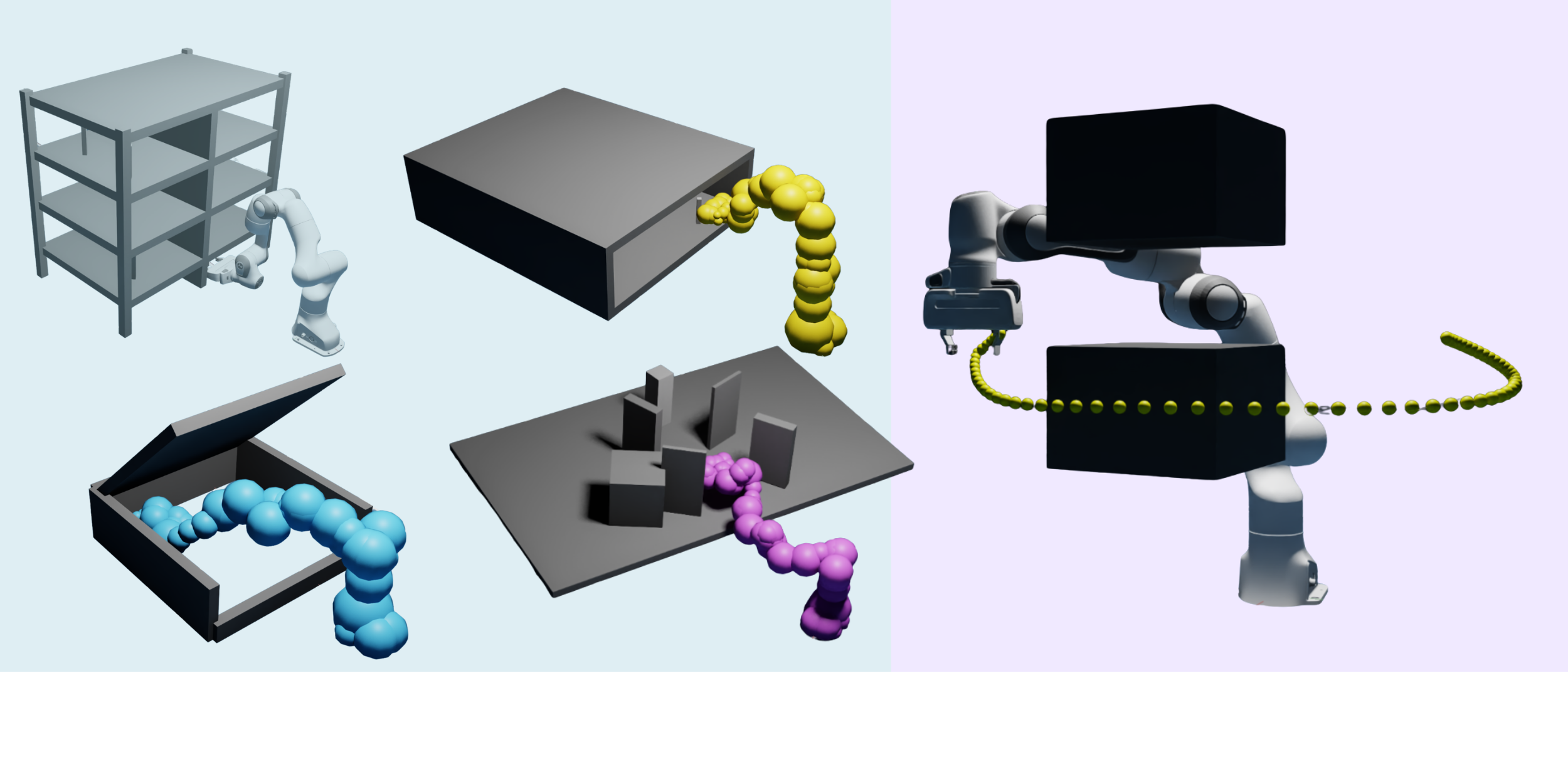}
        \put(14, 39.5){\textbf{Evaluation A}}
        \put(65, 39.5){\textbf{Evaluation B}}
        \put(2, 19){\small Mesh}
        \put(27, 21){\small VSSA}
        \put(0, 2){\small AMAA}
        \put(34.0, 2){\small MorphIt-V}
    \end{overpic}

    \caption{The spherical approximations are evaluated based on the robot's ability to navigate challenging cluttered spaces (refer to \cref{tab:curobo-count} for quantitative results). \textbf{Left)} the robots must reach 100 different goal states within 7 different cluttered environments. The example obstacles are shown in gray and represent a subset of the shelf, box, and tabletop environments used in the benchmark. The robot approximations are shown in successful example goal configurations.
    \textbf{Right)} the robot must traverse between the two cubes with a minimum number of spheres per link used in the morphological approximation.}
    \label{fig:curobo-bench}
\end{figure}

While accurate collision detection is a prerequisite for safe operation, its real-world impact is revealed in how well it supports motion planning in complex environments. For these motion planning evaluations, we select \textsc{MorphIt-V} alongside the two baselines that also achieve zero false negatives in the collision detection analysis (VSSA and AMAA), ensuring that all three methods satisfy the same safety constraint of zero missed collisions based on its balanced performance in the previous collision detection analysis. \cref{fig:curobo-bench} presents two evaluations that assess how the morphological representation impacts a robot's ability to navigate cluttered or tight spaces while reaching its goal state. \cref{tab:curobo-count} shows the results collected from the two evaluations. In Evaluation A, $ns$ is the number of spheres per robot link used for the approximation, while in Evaluation B $ns$ represents the minimum number of spheres per link required by each method to pass through the narrow passage. Additionally, t(s) is the time to generate a successful trajectory, averaged across the 7 evaluation environments.

\begin{table}
% \caption{Results from the collision benchmark evaluation shown on \cref{fig:curobo-bench}. \added{Success rates and solve times are reported as mean $\pm$ 1 std across 25 random seeds, where each seed's value is averaged over the 7 evaluation environments.}}
\caption{Results from the collision benchmark evaluation shown on \cref{fig:curobo-bench}. In Evaluation A, success rates and solve times are reported as mean $\pm$ 1 std across 25 random seeds, where each seed's value is averaged over the 7 evaluation environments. In Evaluation B, we report the minimum sphere count $\alpha$ found and its corresponding solve time.}
\centering
\resizebox{\columnwidth}{!}{%
\begin{tabular}{l|ll|ll|}
\cline{2-5}
                                       & \multicolumn{2}{c|}{\textbf{\begin{tabular}[c]{@{}c@{}}Evaluation A \\ $ns=6$\end{tabular}}}                          & \multicolumn{2}{c|}{\textbf{\begin{tabular}[c]{@{}c@{}}Evaluation B\\ $ns=\alpha$\end{tabular}}} \\ \cline{2-5} 
                                       & \multicolumn{1}{l|}{Success}                  & t(s)                                  & \multicolumn{1}{l|}{\textbf{$\alpha$}} & t(s) \\ \hline
\multicolumn{1}{|l|}{\textbf{VSSA}}     & \multicolumn{1}{l|}{79.14\% $\pm$ 8.41}       & 0.089 $\pm$ 0.012                     & \multicolumn{1}{l|}{16}                & 0.61 \\ \cline{1-1}
\multicolumn{1}{|l|}{\textbf{AMAA}}     & \multicolumn{1}{l|}{87.57\% $\pm$ 5.09}       & 0.074 $\pm$ 0.010                     & \multicolumn{1}{l|}{10}                & 0.28 \\ \cline{1-1}
\multicolumn{1}{|l|}{\textbf{MorphIt-V}} & \multicolumn{1}{l|}{\textbf{91.29\% $\pm$ 3.99}} & \textbf{0.063 $\pm$ 0.006}            & \multicolumn{1}{l|}{\textbf{6}}        & 0.21 \\ \hline
\end{tabular}%
}
\label{tab:curobo-count}
\end{table}

% \begin{table}
% \caption{Results from the collision benchmark evaluation shown on \cref{fig:curobo-bench}, Evaluation A ($ns=6$). Success rates and solve times are reported as mean $\pm$ 1 std across N seeds, averaged over the 7 evaluation environments.}
% \centering
% \begin{tabular}{|l|l|l|}
% \hline
%                  & Success                          & t(s)                       \\ \hline
% \textbf{VSSA}    & 79.14\% $\pm$ 8.41               & 0.089 $\pm$ 0.012          \\ \hline
% \textbf{AMAA}    & 87.57\% $\pm$ 5.09               & 0.074 $\pm$ 0.010          \\ \hline
% \textbf{MorphIt-V} & \textbf{91.29\% $\pm$ 3.99}    & \textbf{0.063 $\pm$ 0.006} \\ \hline
% \end{tabular}
% \label{tab:curobo-count-a}
% \end{table}

% \begin{table}
% \caption{Results from the collision benchmark evaluation shown on \cref{fig:curobo-bench}, Evaluation B ($ns=\alpha$), averaged over the 7 evaluation environments.}
% \centering
% \begin{tabular}{|l|l|l|}
% \hline
%                  & \textbf{$\alpha$} & t(s) \\ \hline
% \textbf{VSSA}    & 16                & 0.61 \\ \hline
% \textbf{AMAA}    & 10                & 0.28 \\ \hline
% \textbf{MorphIt-V} & \textbf{6}      & 0.21 \\ \hline
% \end{tabular}
% \label{tab:curobo-count-b}
% \end{table}

\begin{figure*}[t]
    \centering
    \includegraphics[width=0.99\textwidth , trim=0 125pt 0 0, clip]{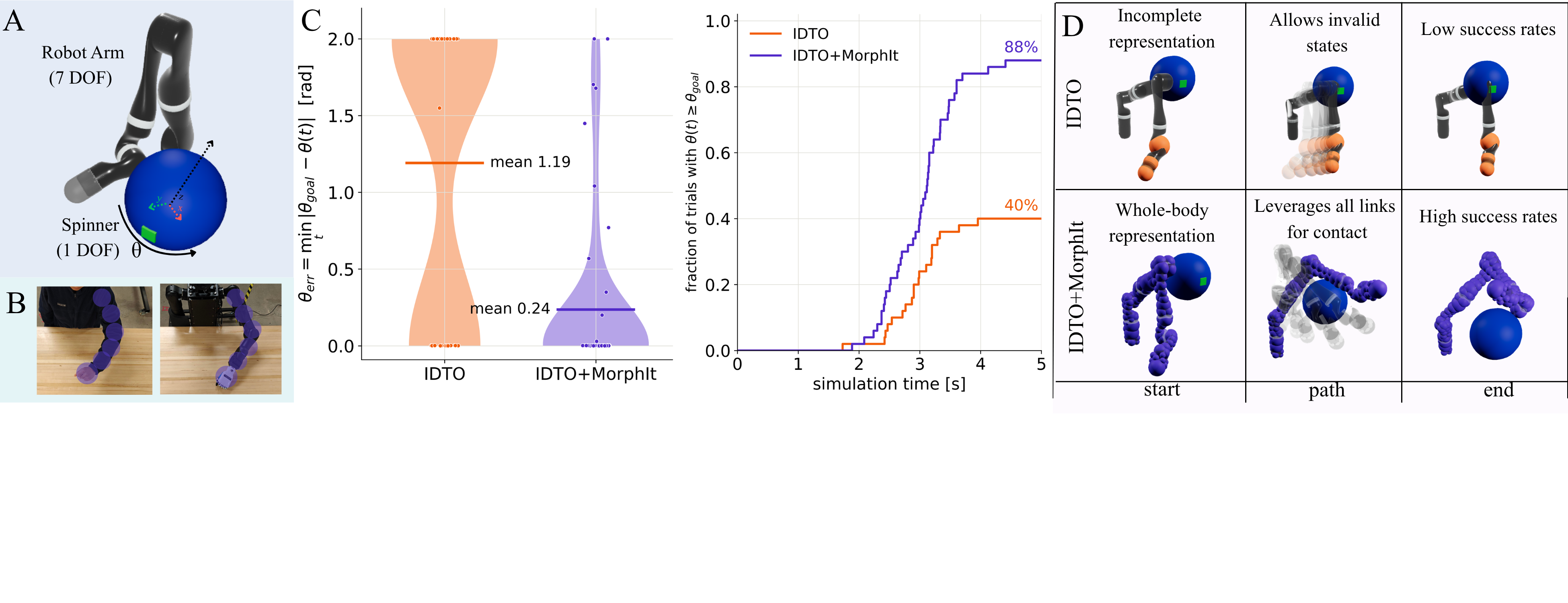} 
    \caption{(A) A 7-DOF Kinova Jaco arm must rotate a \textsl{spinner}, a 1-DOF sphere, to a goal angle $\theta_{goal}$. (B) Real-world analogy demonstrating how humans naturally use their entire body, including elbows, shoulders, and other body parts, for manipulation tasks,  paralleling how our 7-DOF arm can use its elbow joint to rotate the spinner when equipped with whole-body contact modeling. (C) On the left, the distribution of how close each trial brings the spinner to its goal angle, where low values mean the spinner was rotated essentially to the goal and high values mean it was barely moved, with horizontal bars marking the means. On the right, the fraction of trials in which the spinner has reached the goal angle as execution time progresses. (D) Snapshots of a representative trial. With the sparse representation (orange), contact is only possible at a few predetermined wrist points, so the planner sweeps the unmodeled links through the spinner, a physically invalid state that produces no usable contact. With the whole-body representation (purple), every link carries collision geometry, so the planner can form and sustain real contact along the entire arm to drive the spinner.
    Each point in (C) is one deterministic 5\,s closed-loop MPC execution with a 2\,rad goal, with both representations evaluated on the same 50 random spinner placements.}
    \label{fig:idto}
\end{figure*}

The table shows that \textsc{MorphIt-V} achieves up to 12.2 percentage points higher success rates than baseline methods by more accurately representing the robot's geometry, enabling effective navigation through cluttered environments  while preserving the zero-false-negative guarantee required for safe operation. It also produces trajectories the fastest among the three methods, demonstrating that improved geometric fidelity does not come at the cost of planning speed. Evaluation B further demonstrates that \textsc{MorphIt-V} requires up to 62.5\% fewer spheres to successfully navigate a narrow passage compared to VSSA and AMAA methods, resulting in up to 65.6\% reduced computation time. This efficiency stems from \textsc{MorphIt-V}'s volume-prioritized coverage, which produces conservative approximations that conform tightly to the robot's geometry facilitating collision-free navigation through constrained spaces while maintaining computational efficiency. Navigating narrow spaces is essential for robots operating in manufacturing plants, chemistry labs, and aboard space stations where the operational space is limited. In such mission-critical settings, failure to navigate tight environments can lead to task failure, equipment damage, or safety risks. Ensuring reliable performance in these contexts is vital for the success and safety of robots.

\subsection{Results L3: Whole-body Manipulation}
Beyond obstacle avoidance, effective physical interaction represents another essential capability. While cuRobo focuses on obstacle avoidance, IDTO finds trajectories that move target objects from start to goal configurations. For this contact-rich evaluation, we selected \textsc{MorphIt-S} based on its superior contact dynamics performance.

As illustrated in \cref{fig:idto}, when the Jaco arm uses only three spheres at the wrist and fingertip to rotate a spinner, it becomes limited to those contact points, restricting movement and frequently failing. Across the 50 random spinner placements the sparse representation never establishes contact in 29 trials and reaches the goal angle in only 20 of the 50 trials. In contrast, \textsc{MorphIt-S} enables multiple contact points through full-body spherical representation, allowing the arm to leverage its entire geometry. With the whole-body representation the arm reaches the goal in 44 of the 50 trials, driving the spinner with its forearm and elbow when the spinner lies outside the reach of the wrist. Implementation with IDTO demonstrates an $\sim$80\% reduction in mean angular error, from 1.19\,rad to 0.24\,rad, compared to the original sparse algorithm, and the fraction of trials that reach the goal over time shows the same trend (\cref{fig:idto}). This enables more dexterous whole-body interactions, such as using the forearm and elbow to spin the object, analogous to how humans naturally use their entire body for complex manipulation tasks. Critically, the original sparse IDTO allows physically invalid configurations where the robot body passes through the spinner, as it lacks body collision awareness. \textsc{MorphIt-S} addresses this by providing full-body collision awareness, preventing such invalid configurations.

Our results demonstrate that state-of-the-art algorithms achieve significantly richer manipulation behaviors when equipped with whole-body geometric representations. By providing IDTO with \textsc{MorphIt}'s full-body approximation, we unlock implicit manipulation strategies that leverage the robot's entire physical form rather than predetermined contact points. We demonstrate this on a 7-DOF manipulator, and without loss of generality, \textsc{MorphIt}'s approach applies to humanoid, dual-arm manipulator, quadruped and other robot morphologies. The key enabler is accurate spherical representation that leads to physically intelligent behaviors.

\section{Limitations} \label{sec:limitations}
Our evaluation is conducted primarily in simulation, which provides the controlled environment necessary to manipulate morphological representation as the independent variable. Similar real-world validation would present substantial experimental challenges. For example, replicating our collision detection benchmark would require millimeter-precision contact localization across hundreds of scenes and thousands of robot configurations, which is difficult with current perception systems.

Our comparative evaluation focuses on baseline methods (VSSA, AMAA, and cuRoboV1) that generate standard sphere primitives.
While alternative representations exist, including sphere-meshes \citep{thiery2013} and recent neural network-based approaches \citep{yang2024robotsdf, park2019deepsdf, wang2024nerf}, these cannot be directly integrated into standard robotics frameworks relying on URDF specifications, collision checkers like Pinocchio and FCL, motion planners like CuRobo and OMPL, or physics simulators like MuJoCo, PyBullet, and Drake. We acknowledge that neural approaches offer high fidelity and fast per-query inference. However, \textsc{MorphIt}'s explicit sphere representation provides key advantages: analytically differentiable $\mathcal{O}(1)$ distance queries, rotational invariance, and direct compatibility with decades of robotics infrastructure. Furthermore, comparing algorithmic geometric methods to learned representations constitutes a different problem paradigm. Neural approaches require substantial training data collection tailored to each robot embodiment and task context, whereas \textsc{MorphIt} provides one-shot geometric optimization. A comprehensive comparison between these paradigms represents an important but distinct research direction beyond the scope of this work.

\section{Conclusion and Future Work} \label{sec:conclusion}
\textsc{MorphIt} introduces a flexible approach to robot morphology approximation that bridges geometric accuracy and computational efficiency through task-specific adaptation. Our results demonstrate that adapting morphological representations to task constraints improves planning, interaction, and simulation performance. 

While our experiments use approximations pre-computed offline, \textsc{MorphIt}'s computational efficiency suggests an exciting path toward online adaptation. In future systems, robots could dynamically adjust their geometric representations to match current demands: switching from coarse models during long-horizon navigation to fine-grained approximations for constrained manipulation, refining representations during tool use, or reconfiguring models for modular robots as their physical morphology changes. Realizing this vision will require automating weight selection, for example via meta-learning frameworks that tune loss weights based on downstream task performance. As robots operate in increasingly complex and dynamic environments, the ability to rapidly adapt representational complexity to task demands will be a key component in scalable, physically intelligent behavior.

%\section{References}
\bibliographystyle{SageH}
\bibliography{9_ref.bib}

\vspace{-5pt}
\section{Appendix} \label{sec:appendix}

\subsection{Computational Resources} \label{sec:appendix-code}
All experiments were conducted on a workstation equipped with an 
NVIDIA GeForce RTX 3070 GPU (GA104), an Intel Core i9 CPU, and 
16 GB of RAM. The baseline methods and our implementation are 
available as open source.
\footnote{VSSA:\url{https://github.com/CK1201/sphere-set-approximation}}%
\footnote{AMAA:\url{https://github.com/mlund/spheretree}, 
\url{https://github.com/CoMMALab/foam}}%
\footnote{CuRoboV1:\url{https://github.com/NVlabs/curobo}}
\footnote{MorphIt:\url{https://github.com/HIRO-group/MorphIt-1}}

\subsection{Determining limits for number of spheres} \label{sec:appendix-num-spheres}
To determine the number of spheres for approximating each robot link, we referenced CuRobo's widely used library for efficient motion planning. Their manually-packed spherical model of the Panda robot uses an average of 6 spheres per link (range: 2--18).
Moreover, we consider the computational constraints of current systems. CuRobo's self-collision pipeline efficiently processes approximately 20-25 spheres per link on the Panda before performance degradation occurs and the system must switch to a computationally expensive kernel. 
While faster hardware could improve performance, we focus our analysis on CuRobo's current optimal ranges and avoid approximations exceeding 100 spheres per mesh object. More specifically, we tested all methods using sphere counts from 1 to 10 and their squared values. This is because AMAA's tree structure only allows us to set a branch and depth factor rather than a desired number of spheres, whereas VSSA and \textsc{MorphIt} allow the approximation for any exact number of spheres. We calibrate the input parameters, such as number of sample points per mesh, for every algorithm.

\subsection{Adapting IDTO to our task} \label{sec:appendix-idto}
The IDTO formulation relies on a compliant contact model where the normal force is given by $f_n = c(\phi)d(v_n)$, with the stiffness component $c(\phi) = \sigma k \log(1 + \exp(-\phi/\sigma))$ approximating a linear spring of stiffness $k$ in the limit of zero smoothing parameter $\sigma$. The parameter $\sigma$ controls the tradeoff between smoothness and force at a distance, directly impacting both the optimization landscape and the physical realism of the solution. Small $\sigma$ values yield more realistic contact behavior but create steep cost landscapes that prevent convergence, while large $\sigma$ values smooth the optimization problem at the expense of introducing significant force at a distance artifacts. This tradeoff prevented us from initializing the robot end effector at arbitrary start configurations, as the optimizer would either fail to converge with physically realistic $\sigma$ values or produce unrealistic trajectories with overly smooth contact models. To address this limitation, we implemented a trajectory initialization process using inverse kinematics that interpolates between the target object and the closest sphere on the robot at its start configuration. While this closest sphere is often not the final manipulation point and may sometimes be located on the base link or unable to reach the spinner, this approach provides sufficient initial conditions for the optimizer to converge with smaller $\sigma$ values that maintain physical plausibility.

\subsection{Empirical Study of Alternative Primitives} \label{sec:appendix-primitives}

\begin{figure}
    \centering
    \includegraphics[width=0.45\textwidth]{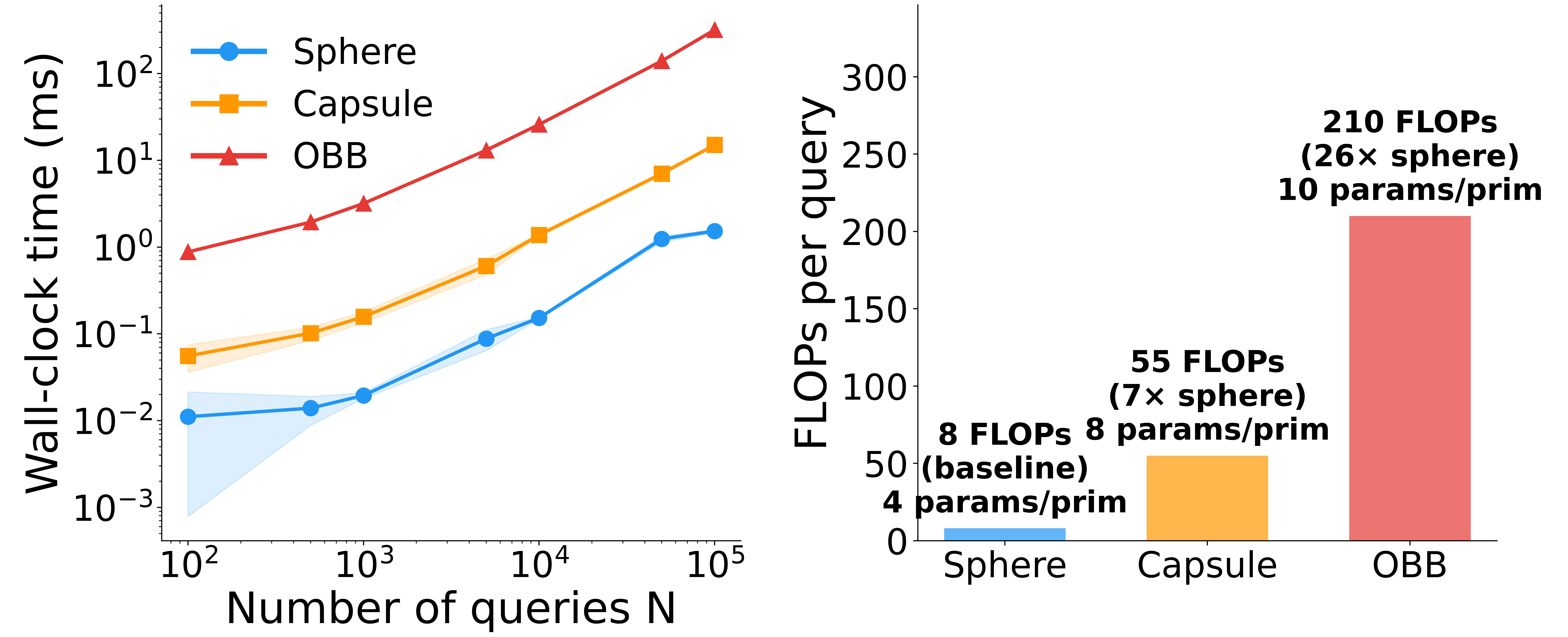} 
    \caption{Distance query benchmark comparing spheres, capsules, 
    and OBBs. Left: wall-clock time versus query count. Wall-clock times are reported as mean $\pm$ 1 std across 10 repeated timing runs.}
    \label{fig:study1}
    \vspace{-5pt}
\end{figure}

\begin{figure}
    \centering
    \includegraphics[width=0.45\textwidth, trim=0 0 0 0, clip]{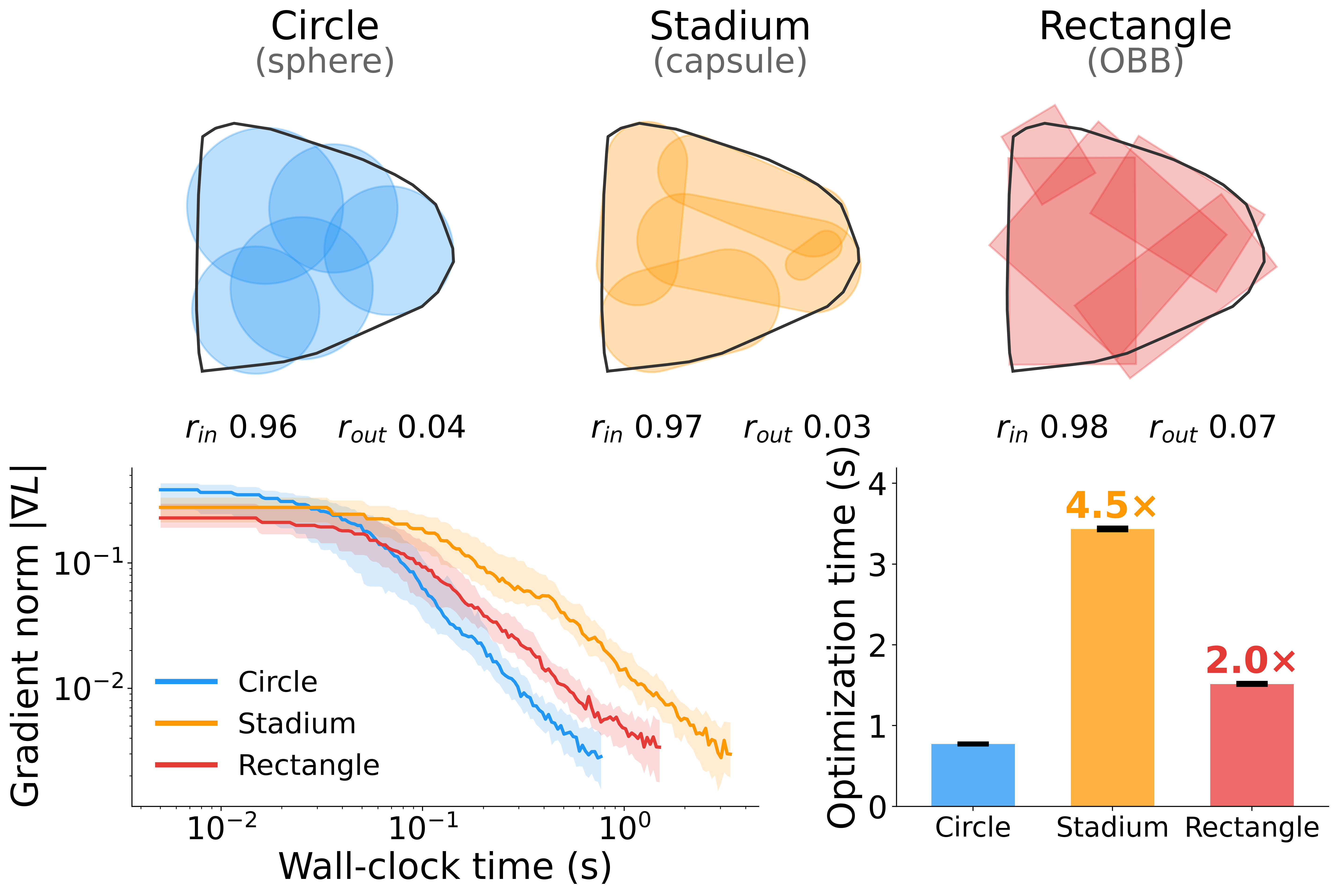} 
    \caption{
    We pack cross-sections of Panda link 0 (the sections of
    \cref{fig:study3}) with circles, stadiums, and rectangles, the 2D
    counterparts of spheres, capsules, and OBBs, chosen for simplicity
    and visual clarity. Under the same packing objective and identical
    optimizer settings, all three types reach comparable quality in the
    2D analogues of our occupancy ratios (top), while circles fit
    fastest, with gradient norms that decay sooner (bottom left) and
    lower total optimization time (bottom right), median with
    interquartile band across 5 sections $\times$ 10 seeds. We expect a
    systematic 3D study to show a similar trend, which we leave to
    future work.}
    \vspace{-3pt}
    \label{fig:study2}
\end{figure}

\begin{figure}
    \centering
    \includegraphics[width=0.45\textwidth, trim=0 0 0 0, clip]{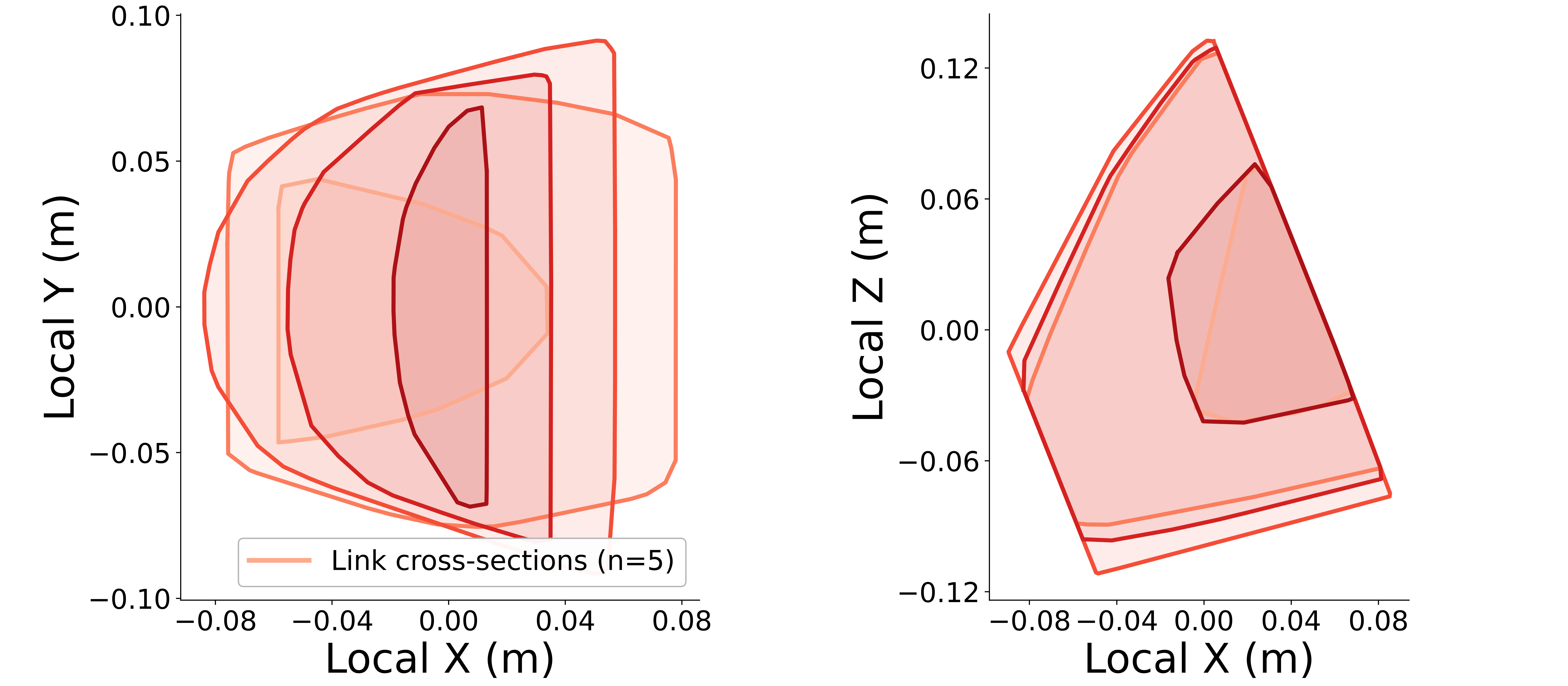} 
    \caption{Cross-sections of the Franka Panda link 0 taken along 
    two orthogonal planes ($n=5$ evenly spaced slices each). The 
    irregular, asymmetric, and inconsistent slice shapes demonstrate 
    that real robot geometry does not conform to any single idealized 
    primitive.}
    \label{fig:study3}
    \vspace{-5pt}
\end{figure}

We present a focused empirical study comparing spheres 
against capsules and oriented bounding boxes (OBBs) across three 
dimensions: distance query efficiency, optimization complexity, 
and geometric compatibility with real robot morphology. This is 
not intended as an exhaustive primitive comparison and we 
encourage further work in this area.

First, spheres are rotationally invariant, reducing 
pairwise distance queries to simple center-to-center Euclidean 
arithmetic ($\sim$8 FLOPs). Capsules require complex 
segment-to-segment branching logic ($\sim$55 FLOPs), and our 
algorithmic vectorized benchmarks confirm that capsule distance queries are 
8.2$\times$ slower than spheres, while OBBs are more than 
174$\times$ slower (see \cref{fig:study1}). We note that this per-query advantage does not by itself imply that spheres are the optimal choice at every operating point. At lower primitive counts, other primitives may in principle be able to better exploit specific link geometries: capsules may more efficiently span long, slender limbs, and cuboids may better approximate box-like torso segments, potentially requiring fewer primitives than an equivalent spherical packing to achieve comparable coverage. However, as shown in \cref{fig:study3}, real robot links, such as Panda Link 0, exhibit irregular, asymmetric cross-sections that deviate substantially from any single idealized primitive. We therefore expect that as the desired reconstruction fidelity increases, more primitives are needed to conform to this irregularity, and that the early efficiency advantage of shape-matched primitives such as capsules or cuboids diminishes. We emphasize that this is a conjecture rather than a measured result. We emphasize that this is a conjecture rather than a measured result. We acknowledge that establishing the primitive count at which spheres overtake capsules or cuboids for a given fidelity target would require a systematic fidelity-versus-count study for each primitive type, which we leave to future work. What our benchmarks do establish is that whenever a comparable number of primitives is required, spheres are substantially cheaper per query. Extending \textsc{MorphIt} to support a mixture of primitive types, and identifying the operating regimes in which each type is preferable, is a promising direction for future work.

Second, spheres are also cheaper to optimize. A sphere has
4 parameters and no rotational degrees of freedom, against 8 for
capsules and 10 for OBBs. We measure the effect by
packing 2D cross-sections of Panda link 0 with $n=5$ primitives of each
type under a shared objective and identical optimizer settings
(\cref{fig:study2}). All three types reach comparable packing quality,
while circles complete the fit 4.5$\times$ faster than capsules and
2.0$\times$ faster than rectangles. Although packings are precomputed
offline, fitting time remains relevant for method development and for
the online adaptation scenarios discussed in \cref{sec:conclusion}.

\subsection{Formulation of Baseline Methods}
\label{app:baseline_objectives}
To clarify the relationship between the baseline objectives and our 
framework, we draw analogies between the fixed objectives of AMAA and VSSA 
and the loss terms defined in Table~1. While no exact correspondence exists, 
this mapping illustrates which criteria each baseline implicitly 
prioritizes.

\subsubsection{Variational Sphere Set Approximation (VSSA)}
VSSA applies iterative Lloyd clustering to find a sphere set 
$\mathcal{S} = \{S_1, \dots, S_{n_s}\}$ that minimizes the Sphere Outside 
Volume (SOV), defined as the total volume of spheres that protrudes beyond 
the object surface:
\begin{equation}
    E_{\text{VSSA}}(X, \mathcal{S}) = \sum_{i=1}^{n_s} \iiint_{y \in S_i} 
    \mathbf{1}[y \notin X] \, dy \label{eq:vssa}
\end{equation}
where $\mathbf{1}[y \notin X]$ is an indicator returning $1$ if point 
$y$ lies outside the object $X$ and $0$ otherwise. Mapped onto our framework 
(Table~1), VSSA optimizes an objective similar to 
$\mathcal{L}_{\text{boundary}}$ alone, with $w_{\text{overlap}} = 
w_{\text{containment}} = w_{\text{surface}} = w_{\text{SQEM}} = 0$ by 
construction. The original authors note that neglecting interior overlap is 
a deliberate design choice that provides algorithmic flexibility.

\subsubsection{Adaptive Medial-Axis Approximation (AMAA)}
AMAA generates spheres by placing them at internal Voronoi vertices of 
the object's approximate medial axis, where each sphere's radius equals the 
distance to the nearest surface point. The bounding error $e$ for a sphere 
centered at Voronoi vertex $v$ with radius $r$ relative to the closest surface 
point $q$ is:
\begin{equation}
    e = r - \|q - v\| \label{eq:amaa}
\end{equation}
AMAA then greedily minimizes the maximum $e$ across all spheres through 
iterative merging and bursting operations, reducing the sphere count while 
bounding worst-case protrusion. In the context of our framework, this 
corresponds to optimizing an analogue of $\mathcal{L}_{\text{surface}}$ alone, 
through a sequential greedy heuristic rather than a continuous global 
objective. As a result, the method does not expose tunable weights over 
competing geometric criteria. It is specialized for minimizing worst-case 
surface protrusion.

\subsubsection{cuRoboV1} \label{app:curobo}
CuRobo approximates a mesh with $N$ spheres via a voxelization-based procedure. Given the mesh and the requested sphere count $N$, the algorithm first computes a uniform voxel pitch $p = (V / N)^{1/3}$, where $V$ is the mesh's enclosed volume, and assigns each candidate sphere a radius of $p/2$. The mesh is voxelized at this pitch using ray casting, yielding a grid of candidate sphere centers. Each candidate is retained only if the signed distance from its center to the mesh surface exceeds $p/2$, guaranteeing that the corresponding sphere fits strictly inside the volume. When this inside-test discards all candidates, the algorithm falls back to even surface sampling, which is a common failure mode at low $N$ on thin or concave geometries where no interior point is deeper than $p/2$ from the surface. In this case, the algorithm returns $N$ evenly-spaced points on the mesh boundary assigned a user-specified \texttt{surface\_sphere\_radius}. The output is therefore a strictly-inside volumetric packing when the geometry permits, and an inflated surface approximation otherwise. The only user-exposed parameters are $N$, the fit type, and the fallback radius.

\subsection{Ablation Study: Adaptive Density Control}\label{app:ablation}
\begin{figure}[th]
    \centering
    \includegraphics[width=0.5\textwidth, trim=0 0 0 0, clip]{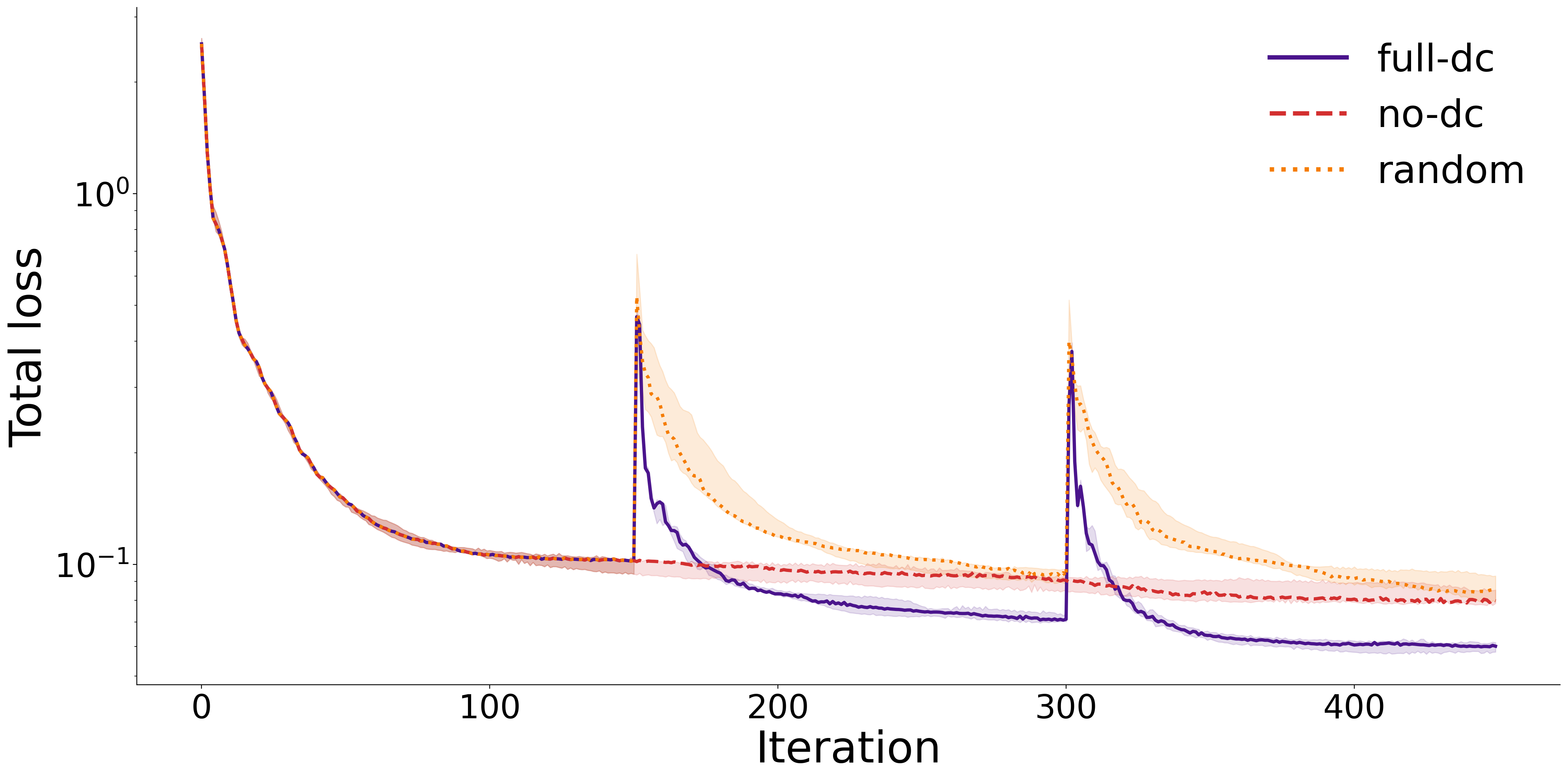} 
    \caption{Total loss over training iterations for the three ADC conditions 
    (\textit{full-dc}, \textit{no-dc}, \textit{random}) at $n{=}64$ spheres, 
    averaged over 20 runs (shaded regions denote $\pm1$ standard deviation). 
    Vertical spikes correspond to ADC events, where spheres are culled and 
    reseeded; \textit{full-dc} recovers rapidly and converges to a lower final 
    loss than either ablated condition.}
    \vspace{-5pt}
    \label{fig:loss_trajectory}
\end{figure}

We evaluate the contribution of adaptive density control by comparing 
three conditions across four Panda collision meshes, four links, and five random 
seeds (20 runs per condition): \textit{full-dc} (our complete method), 
\textit{no-dc} (adaptive density control disabled throughout optimization), and \textit{random} 
(the density control schedule is preserved, but cull selection and reseed positions are randomized). 
Results are summarized in Table~\ref{tab:adc_ablation} and 
Fig.~\ref{fig:loss_trajectory}.
\textit{full-dc} consistently achieves lower surface distance, boundary error, 
and total loss than both ablated conditions across all sphere budgets. The elevated variance of \textit{random} confirms that coverage-guided 
redistribution, rather than the density control schedule itself, is the operative mechanism.

\begin{table}[h]
\centering
\caption{Ablation of adaptive density control across two sphere budgets,
evaluated on four FR3 collision meshes over 20 runs per condition. Values are mean $\pm$ 1 std.}
\resizebox{\columnwidth}{!}{%
\begin{tabular}{lccccc}
\toprule
\multicolumn{6}{c}{Number of spheres = 6} \\
\midrule
Condition & $d_{avg}$ (mm) & $d_{max}$ (mm) & $r_{out}$ & $r_{union}$ & Loss \\
\midrule
full-dc & $4.68 \pm 1.33$ & $18.67 \pm 4.08$ & $0.202 \pm 0.086$ & $1.157 \pm 0.084$ & $0.739 \pm 0.354$ \\
no-dc   & $5.12 \pm 1.46$ & $19.64 \pm 4.71$ & $0.224 \pm 0.091$ & $1.176 \pm 0.088$ & $0.847 \pm 0.409$ \\
random  & $5.77 \pm 1.66$ & $21.47 \pm 4.93$ & $0.263 \pm 0.145$ & $1.210 \pm 0.140$ & $1.017 \pm 0.473$ \\
\midrule
\multicolumn{6}{c}{Number of spheres = 64} \\
\midrule
full-dc & $1.34 \pm 0.29$ & $8.19 \pm 2.30$ & $0.041 \pm 0.012$ & $1.017 \pm 0.012$ & $0.098 \pm 0.029$ \\
no-dc   & $1.53 \pm 0.40$ & $9.62 \pm 2.74$ & $0.050 \pm 0.014$ & $1.024 \pm 0.014$ & $0.130 \pm 0.051$ \\
random  & $1.63 \pm 0.43$ & $9.47 \pm 2.83$ & $0.052 \pm 0.010$ & $1.026 \pm 0.010$ & $0.140 \pm 0.056$ \\
\bottomrule
\end{tabular}}
\label{tab:adc_ablation}
\end{table}

\end{document}